\documentclass[11pt]{article}

\usepackage[preprint]{acl}

\usepackage{times}
\usepackage{latexsym}
\usepackage{booktabs}
\usepackage[T1]{fontenc}

\usepackage[utf8]{inputenc}

\usepackage{microtype}

\usepackage{inconsolata}

\usepackage{graphicx}
\usepackage{adjustbox}
\usepackage{multirow}
\usepackage{colortbl}
\usepackage[dvipsnames]{xcolor}
\usepackage{rotating}
\usepackage{tikz}
\usetikzlibrary{positioning,calc,arrows.meta,shapes.geometric}
\usepackage{palette}
\usepackage{amsmath}
\usepackage{amssymb}
\usepackage[most]{tcolorbox}

\usepackage[capitalize]{cleveref}
\crefname{figure}{Fig.\@}{Figs.\@}
\Crefname{figure}{Fig.\@}{Figs.\@}
\Crefname{section}{Sec.\@}{Secs.\@}
\Crefname{appendix}{App.\@}{Apps.\@}

\newcommand{\Gi}{\mathcal{G}_i}
\newcommand{\Gu}{\mathcal{G}_t}

\newcommand{\Gpre}{\mathcal{G}}
\newcommand{\Gri}{\mathcal{G}^r_i}
\newcommand{\Grt}{\mathcal{G}^r_t}
\newcommand{\Grpre}{\mathcal{G}^r}
\newcommand{\Vi}{\mathcal{V}_i}
\newcommand{\Vu}{\mathcal{V}_t}
\newcommand{\Rin}{\mathcal{R}_i}
\newcommand{\Ru}{\mathcal{R}_t}
\newcommand{\Ei}{\mathcal{E}_i}
\newcommand{\Eu}{\mathcal{E}_t}
\newcommand{\hth}{\text{h2h}}
\newcommand{\tth}{\text{t2h}}
\newcommand{\ttt}{\text{t2t}}
\newcommand{\htt}{\text{h2t}}
\newcommand{\TRIXnoiter}{TRIX$_{\text{noiter}}$}
\newcommand{\ULTRArand}{ULTRA$_{\text{rand}}$}

\newcommand{\reftop}[1]{\cref{#1}~(top)}
\newcommand{\refbot}[1]{\cref{#1}~(bottom)}

\newcommand{\halflingmark}{\protect\raisebox{-0.5pt}{\protect\includegraphics[height=10pt]{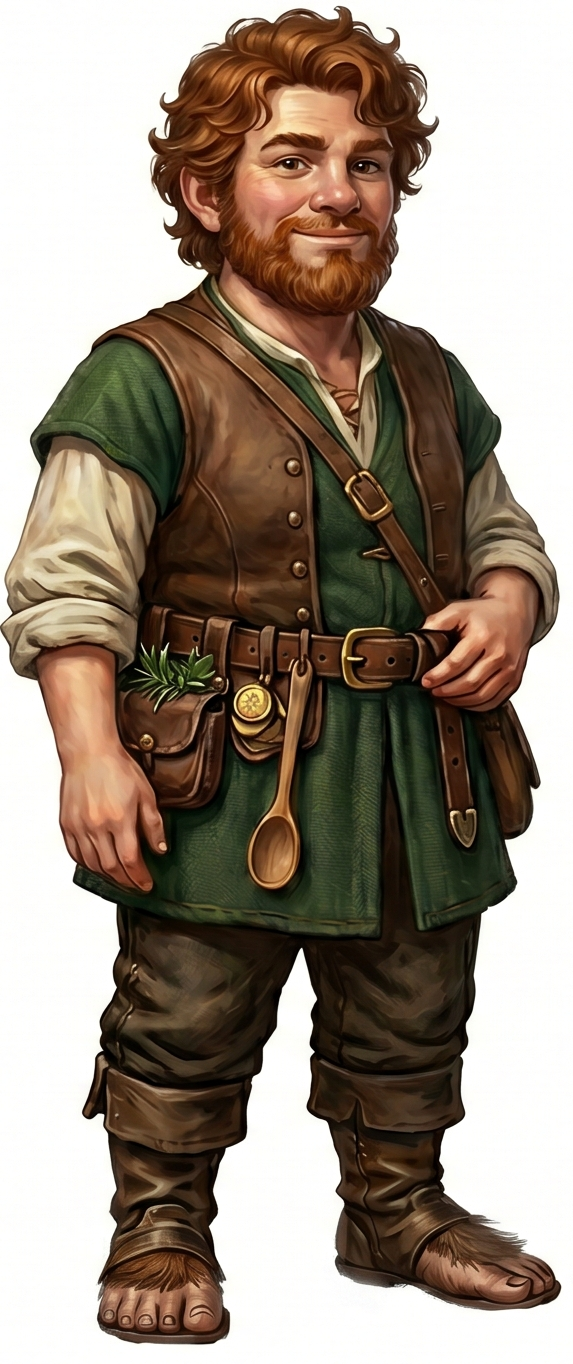}}}

\newtcolorbox[auto counter]%
  {takeawaybox}[1][]{colframe=petroil2,
    colback=lacamlilac!04!white,
    colbacktitle=petroil2,
    fonttitle=\bfseries,
    arc=0pt, outer arc=0pt,
    enhanced,
    attach boxed title to top left={yshift=-1pt},
    boxed title style={arc=0pt, outer arc=0pt},
    #1}

\definecolor{outrageousorange}{rgb}{1.0, 0.43, 0.29}
\hypersetup{
colorlinks=true,linkcolor=Orchid,citecolor=outrageousorange!90
}

\title{{Half a Link can Be Enough to Predict a Whole Link:\\Understanding Generalization in Knowledge Graph Foundation Models}}

\author{%
  \textbf{Cosimo Gregucci}\textsuperscript{1},
  \textbf{Obaidah Theeb}\textsuperscript{1},
  \textbf{Daniel Hernández}\textsuperscript{1},\\
  \textbf{Antonio Vergari}\textsuperscript{3,\halflingmark},
  \textbf{Steffen Staab}\textsuperscript{1,2,\halflingmark}
\\
  {\normalfont\textsuperscript{1}Institute for AI, University of Stuttgart,
  \textsuperscript{2}University of Southampton,}\\
  {\normalfont\textsuperscript{3}University of Edinburgh}
\\
  {\normalfont\texttt{cosimo.gregucci@ki.uni-stuttgart.de}}%
}

\renewcommand{\paragraph}[1]{\vspace{2pt}\textbf{#1}}

\begin{document}
\maketitle
\renewcommand{\thefootnote}{\halflingmark}%
\footnotetext{Shared supervision.}%
\renewcommand{\thefootnote}{\arabic{footnote}}%
\setcounter{footnote}{0}%
\begin{abstract}

Knowledge graph (KG) foundation models (KGFMs) are zero-shot generalizers: trained once, they can predict links on unseen graphs without retraining. 
However, understanding when and how they can robustly generalize across KGs is still an open question.
In this paper, we shed some light on their generalization mechanisms highlighting how their performance on unseen KGs is not uniform when it comes to partially seen links, which we call \textit{half-links}.
In fact, we show that to predict a test triple $(h,r,t)$ it might suffice in practice to have observed the half-link $(h,r)$ or  
$(r,t)$ in the inference graph.
This yields a taxonomy of four scenarios when combinations of these half-links are observed or not. 
In a rigorous stratified analysis over these scenarios, we reveal that 
SoTA KGFMs use seen half links for predictions, while unseen half-links pose different challenges.
As such, our finer-grained taxonomy can be a diagnostic protocol for robust KGFM generalization and highlights where novel KGFMs can improve.

\end{abstract}

\section{Introduction}

\newsavebox{\tipCompBox}\sbox{\tipCompBox}{\tikz[baseline=-0.5ex]{\fill[bgrey1] (0,0) -- (3pt,1.5pt) -- (3pt,-1.5pt) -- cycle;}}%
\newsavebox{\tipHasBox}\sbox{\tipHasBox}{\tikz[baseline=-0.5ex]{\fill[bgrey1] (-2pt,-2pt) rectangle (2pt,2pt);}}%
\begin{figure*}[!t]
  \centering
  \includegraphics[width=\linewidth]{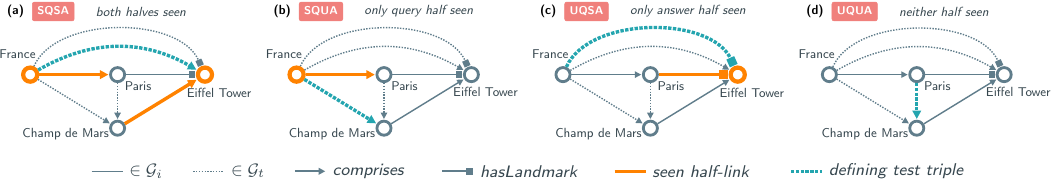}
\caption{\textbf{The inference graph supports test triples to fundamentally different scenarios,}
from both halves seen (SQSA), to one (SQUA, UQSA), to neither (UQUA).
On a shared graph, the inference graph $\Gi$ (solid) and the test graph $\Gu$ (dotted), each scenario refers to a single test triple, the thick, dotted \textcolor{petroil2}{\textbf{teal}} edge; the scenario it falls into depends on which of its halves are \emph{seen}, that is, matched by a thick \textcolor{gold2}{\textbf{orange}} half-link in $\Gi$. The other candidate triples stay thin and gray.
Arrowhead shape encodes the relation name (\usebox{\tipCompBox}~\emph{comprises}, \usebox{\tipHasBox}~\emph{hasLandmark}).}
\label{fig:emnlp-quadrants}
\end{figure*}

Knowledge graphs (KGs) encode factual knowledge as a set of $(h,r,t)$ triples, each linking a head entity $h$ to a tail entity $t$ through a relation $r$.
KGs underpin applications such as virtual assistants and 
recommendation~\citep{DBLP:conf/sigmod/IlyasRKPQS22,DBLP:conf/kdd/Dong18} and are used to store incomplete data~\citep{DBLP:journals/csur/HoganBCdMGKGNNN21}. 
\emph{Link prediction} is the task of
predicting unseen triples by answering 
queries of the form $(h,r,?)$, i.e., finding which entities are related to a head $h$ through the relation name $r$.

KG foundation models (KGFMs) make link prediction zero-shot: trained once on a collection of source graphs, they predict links on unseen graphs without retraining~\citep{DBLP:conf/iclr/0001YM0Z24,DBLP:conf/icml/LeeCW23}.
In the \textit{inductive transfer}\footnote{Also called fully-inductive in \citet{DBLP:conf/icde/GengCPCJZC23}.} setting we study (\cref{sec:found}), a KGFM is pre-trained once and then, at test time, conditioned on a new inference graph $\Gi$ with edge set $\Ei$, whose entities and relation names did not appear during pre-training. 
Then, it answers queries $(h,r,?)$ over a held-out edge set $\Eu$ that is disjoint from $\Ei$, but involving entities and relations in $\Ei$, ranking every candidate entity $t$.

Several works measure the generalization power of KGFMs in terms of aggregated raw performance, i.e., the ability to predict test triples according to ranking metrics, such as the mean reciprocal rank (MRR) averaged over several datasets~\citep{DBLP:conf/iclr/0001YM0Z24}.
At the same time, other works questioned the ability of KGFMs, and link predictors in general \citep{DBLP:conf/nips/ZhuZXT21, DBLP:conf/icml/TrouillonWRGB16,DBLP:conf/iclr/SunDNT19,DBLP:conf/www/GregucciN0S23,DBLP:conf/nips/LoconteMPV23}, to robustly generalize, also in the context of complex query answering \citep{DBLP:conf/icml/GregucciX0LMSV25}. 
This is done by looking at the ability of link predictors to \textit{memorize whole links} \citep{DBLP:conf/nips/NickelJT14}, and to guarantee that training links do not leak into test KGs \citep{DBLP:conf/emnlp/ArunKNXKVS25,DBLP:conf/acl-cvsc/ToutanovaC15}. 
In this paper, we offer a novel and finer-grained perspective on the generalization power of KGFMs by highlighting how their performance stratifies differently when predicting links for which partial information, which we call \textit{half-links}, has been observed.

Predicting Eiffel Tower for the query $(\text{France},\emph{comprises}, ?)$, the (thick) teal-colored test triple in \cref{fig:emnlp-quadrants}(a), relies on evidence already in $\Gi$: France is a source entity for \emph{comprises} and Eiffel Tower a target one, each seen through a (thick) orange half-link in $\Gi$.
Predicting Champ de Mars for $(\text{Paris}, \emph{comprises}, ?)$, the (thick) teal-colored test triple in \cref{fig:emnlp-quadrants}(d), has no such evidence in $\Gi$: Paris is not a
\emph{comprises}-source and Champ de Mars not a \emph{comprises}-target, so neither half is seen and no half-link in $\Gi$ turns orange, and the model must generalize beyond the direct evidence in $\Gi$.
We argue that the two scenarios above pose different generalization challenges to KGFMs and hence yield different performance.
We formalize our argument by tracing a \textbf{half-link taxonomy}.
We decompose each test triple $(h,r,t)$ into a \textbf{query half} $(h,r,?)$ and an \textbf{answer half} $(?,r,t)$, and call a half \textbf{seen} when $\Gi$ contains a \emph{witness} for it, $(h,r,e)\in\Ei$ for the query half or $(e,r,t)\in\Ei$ for the answer half, and \textbf{unseen} otherwise; an unseen half is precisely where the model must generalize beyond the direct evidence in $\Gi$.
Crossing the two outcomes yields four \textbf{scenarios}, see \cref{sec:beyond}.

\colorlet{rcomp}{gold2}        %
\colorlet{rhas}{petroil2}      %

\tikzset{
  vcol/.style     = {draw=bgrey3, circle, inner sep=0pt, minimum size=10pt,
                      line width=2pt, draw opacity=0.5},
  vcolW/.style    = {vcol, minimum size=8pt, line width=4pt, draw opacity=1},
  vlbl/.style     = {bgrey4, font=\tiny\sffamily, inner sep=1pt, text opacity=0.5},
  vlblIt/.style   = {bgrey4, font=\tiny\sffamily\itshape},
  eobs/.style     = {-stealth, bgrey2, line width=2.8pt},
  ecomp/.style    = {-stealth, rcomp,  line width=2.8pt},
  ehas/.style     = {-stealth, rhas,   line width=2.8pt},
  etst/.style     = {densely dotted, -stealth, line width=1.4pt},
  ecompG/.style   = {ecomp, opacity=0.5, line width=1.2pt},
  ehasG/.style    = {ehas,  opacity=0.5, line width=1.2pt},
  etstG/.style    = {etst,  opacity=0.5},
  qlabBig/.style  = {draw=tomato2, fill=tomato2, inner sep=3.5pt, rounded corners=1.5pt},
  qlabSmall/.style= {draw=tomato0, fill=tomato0, inner sep=3.5pt, rounded corners=1.5pt},
}

Under this finer-grained view, the four scenarios pose genuinely different challenges: averaged over 51 zero-shot benchmarks, the strongest exceeds the weakest by about $0.4$ MRR, a gap as large as the aggregate itself.
Across the benchmarks they are unevenly present: the both-halves-seen scenario dominates and the all-unseen scenario is under $10\%$ on average
(\cref{sec:proportions}).
\textbf{\textit{Half a link can be enough: for example, with only the answer half seen, models reach up to 1.0 MRR on NELLInductive v1}}~\citep{DBLP:conf/icml/TeruDH20}.
This is not leakage: $\Ei$ and $\Eu$ are disjoint, so the answer-half witness $(e,r,t)\in\Ei$ is never the test triple $(h,r,t)$.
The seen-answer advantage survives the node-degree and relation-cardinality confounders previously discussed in \citet{DBLP:conf/uai/MohamedPKA20, DBLP:conf/nips/BordesUGWY13} (\cref{sec:arch}).
Moreover, in the unseen answer-halves scenarios,
ULTRA~\citep{DBLP:conf/iclr/0001YM0Z24} and MOTIF~\citep{DBLP:conf/icml/HuangBBC0RO25} are tied in overall MRR at $0.358$ and $0.359$, yet MOTIF clearly leads ULTRA there, $0.267$ to $0.245$, and almost matches TRIX~\citep{DBLP:conf/log/ZhangBGR24}, the overall-best model (\cref{tab:app_kgfm_perdataset}, \cref{sec:relgraph,sec:results}).

\paragraph{Contributions.}
After establishing the KGFM setting (\cref{sec:found}) and what it means to generalize beyond the inference graph (\cref{sec:beyond}),
(\textbf{C1}) we introduce the half-link taxonomy
and its characterization in the relation graph $\Gri$ that any GNN-based KGFM uses (\cref{sec:beyond}).
(\textbf{C2}) We audit the 57 benchmarks of~\citet{DBLP:conf/iclr/0001YM0Z24} and find that the scenario composition is an artefact of split construction, with the both-halves-seen scenario dominating and the all-unseen scenario under $10\%$ on average (\cref{sec:proportions}).
(\textbf{C3}) We disentangle what drives KGFM performance on answer-seen scenarios and on the all-unseen scenario: a frozen baseline, \ULTRArand, with encoders fixed at random initialization, shows the seen-answer advantage is already architectural, and a relation-graph comparison shows that a more expressive relation graph does not always improve the all-unseen scenario (\cref{sec:arch,sec:relgraph}).
(\textbf{C4}) We re-evaluate ULTRA, MOTIF, and TRIX over 51 zero-shot benchmarks under finer-grained reporting, showing that no KGFM is best on every scenario and that fine-tuning flips the
ranking on the all-unseen scenario (\cref{sec:results}).

\section{KGs \& KGFM for link prediction}\label{sec:found}

\paragraph{Knowledge graphs.}
A knowledge graph (KG) is a tuple $\mathcal{G} = (\mathcal{V}, \mathcal{R}, \mathcal{E})$, where $\mathcal{V}$ is a finite set of entities, $\mathcal{R}$ is a finite set of relation names, and $\mathcal{E} \subseteq \mathcal{V} \times \mathcal{R} \times \mathcal{V}$ is a set of triples. Each triple $(h, r, t) \in \mathcal{E}$ links a head entity $h$ to a tail entity $t$ through a relation name $r$.

\paragraph{Link prediction.}
Given a head entity $h$ and relation name $r$, link prediction ranks all candidate entities $e \in \mathcal{V}$ by the predicted plausibility that $(h, r, e)$ holds, so that the correct but unobserved target $t$ ranks as high as possible.
Under the \textit{filtered} protocol~\citep{DBLP:conf/nips/BordesUGWY13}, letting $s(h,r,e)\in\mathbb{R}$ be the model's score for candidate $e$ and $\mathcal{F}_{h,r,t} = \{t'\in\mathcal{V}: t'\neq t,\,(h,r,t')\in\mathcal{E}\}$ the set of other known true candidates, the rank of the target $t$, denoted as $\operatorname{rank}(h,r,t)$ is:
\begin{align}
    &1 + \bigl|\bigl\{e\in\mathcal{V}\setminus\mathcal{F}_{h,r,t} : s(h,r,e) > s(h,r,t)\bigr\}\bigr|. \notag
\end{align}
Commonly both tail prediction $(h,r,?)$ and head prediction $(?,r,t)$ are evaluated for each test triple, and performance metrics are averaged over both directions.
Performance is reported as mean reciprocal rank (MRR), the average of $1/\operatorname{rank}$ over all (triple, direction) pairs, and Hits@$k$ (H@$k$), the fraction of predictions with rank~$\leq k$.

\paragraph{KG foundation models for link prediction.}
A KG foundation model (KGFM) is pre-trained on a collection of source graphs and evaluated zero-shot on a previously unseen target KG.
We focus on the \textit{inductive transfer} setting,
 in which both entities and relation names are new at inference time.
The target KG is split into the \textit{inference graph} $\Gi = (\Vi, \Rin, \Ei)$, available at inference time, and the \textit{test graph} $\Gu = (\Vu, \Ru, \Eu)$, whose triples must be predicted.
The test entities and relation names occur in the inference graph ($\Vu \subseteq \Vi$, $\Ru \subseteq \Rin$), and the two edge sets are disjoint ($\Ei \cap \Eu = \emptyset$).
The model receives only $\Gi$ and must rank target entities for each test triple $(h, r, t) \in \Eu$.

\begin{figure}[t]
\centering
\begin{minipage}[c]{0.48\linewidth}
\centering
\includegraphics[width=\linewidth]{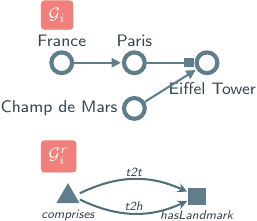}
\end{minipage}\hfill
\begin{minipage}[c]{0.47\linewidth}
\caption{\textbf{Example inference graph $\mathcal{G}_i$ (top) and its ULTRA-derived
relation graph $\mathcal{G}^{r}_i$ (bottom).}
For visual simplification, we omit (i)~\textit{h2t} edges, since they
are the reverse of \textit{t2h}, and (ii)~trivial \textit{h2h} and
\textit{t2t} self-loops.}
\label{fig:rel-graph}
\label{fig:rel-graph:top}
\label{fig:rel-graph:bot}
\end{minipage}
\end{figure}

GNN-based KGFMs use a two-encoder pipeline in which relation representations condition entity-level reasoning.
Given $\Gpre$, they first construct its relation graph $\Grpre$ — with relation names as nodes and co-occurrence motifs as directed edges~\citep{DBLP:conf/icml/LeeCW23, DBLP:conf/iclr/0001YM0Z24}.
For example, following~\citet{DBLP:conf/iclr/0001YM0Z24}, a \textit{t2t} co-occurrence indicates that two relation names share a tail entity, while a \textit{t2h} co-occurrence indicates that the tail of one is the head of the other. In the inference graph in \reftop{fig:rel-graph}, \textit{comprises} and \textit{hasLandmark} share the tail Eiffel Tower, and the tail of \textit{comprises} (Paris) is the head of \textit{hasLandmark}, so the relation graph in \refbot{fig:rel-graph} links them by both a \textit{t2t} and a \textit{t2h} edge.
A \textit{relation encoder} then runs message-passing on $\Grpre$ to produce %
relation embeddings, which feed into an \textit{entity encoder} that runs message-passing on $\Gpre$ to compute entity representations~\citep{DBLP:conf/nips/ZhuZXT21,DBLP:conf/nips/ZhuYGX0G023,DBLP:conf/www/ZhangY22}.

At inference, both encoders run message-passing on the new $\Gi$ (\reftop{fig:rel-graph}) and its derived $\Gri$ (\refbot{fig:rel-graph}), computing fresh entity and relation embeddings specific to the unseen KG; these are then used to score each test triple $(h,r,t)\in\Eu$.
Because neither encoder learns entity- or relation-specific parameters, this process generalizes zero-shot to any unseen KG with arbitrary vocabularies.

We consider three GNN-based KGFMs that instantiate this framework. ULTRA~\citep{DBLP:conf/iclr/0001YM0Z24}
builds the relation graph from binary co-occurrence motifs between relation
names. MOTIF~\citep{DBLP:conf/icml/HuangBBC0RO25} extends it to higher-order ($k \geq 3$) co-occurrence patterns. TRIX~\citep{DBLP:conf/log/ZhangBGR24}
augments the binary relation graph in ULTRA by recording which entities participate
in each co-occurrence, and additionally couples the relation and entity
encoders through iterative updates.
Both MOTIF and TRIX are strictly more expressive than ULTRA, extending it along orthogonal axes: the former through motif length, the latter through per-entity granularity.
We exclude FLOCK~\citep{DBLP:journals/corr/abs-2510-01510}, which replaces deterministic message-passing with probabilistic random-walk ensembles, 
as it requires substantially higher computational cost while achieving comparable zero-shot performance. See \cref{sec:app-flock} for a detailed cost-performance analysis.
\begin{figure*}[!t]
  \centering
  \includegraphics[width=\linewidth]{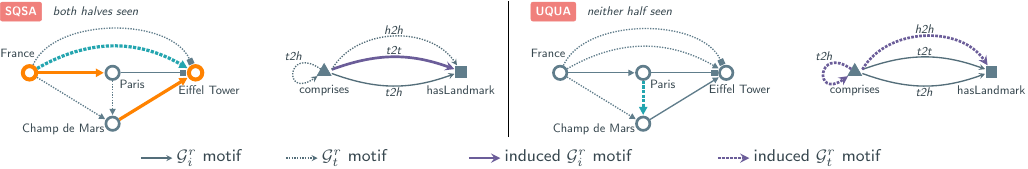}
\caption{\textbf{UQUA test triples can map to potentially missing relation-graph links.}
We draw the co-occurrence motifs each test triple would induce, solid if in $\Gri$ and dotted if in $\Grt$; those induced by the specific test triple (teal, left graph of each pair) are thick \textcolor{lacamlilac}{\textbf{purple}}. \textbf{SQSA} (left): the induced \textit{t2t} between \emph{comprises} and \emph{hasLandmark} is present (purple solid), since by construction the answer entity has an incoming \emph{comprises} edge. \textbf{UQUA} (right): the induced motifs might be missing (purple dotted), recoverable only through incidental co-occurrences (see \cref{sec:relgraph}).}
\label{fig:relimpact}
\end{figure*}

\section{What is generalizing beyond the inference graph?}\label{sec:beyond}
Prior work questions whether KGFMs, and link predictors in general~\citep{DBLP:conf/nips/ZhuZXT21,DBLP:conf/icml/TrouillonWRGB16,DBLP:conf/iclr/SunDNT19}, robustly generalize rather than \emph{memorize} links~\citep{DBLP:conf/nips/NickelJT14}, also in the context of complex query answering~\citep{DBLP:conf/icml/GregucciX0LMSV25}. The held-out edge set $\Eu$ is kept disjoint from $\Ei$, so that no test triple leaks into the inference graph~\citep{DBLP:conf/emnlp/ArunKNXKVS25,DBLP:conf/acl-cvsc/ToutanovaC15}. This guarantee applies at the granularity of the whole triple. Yet even when $(h,r,t)\notin\Ei$, $\Ei$ may still contain \emph{partial information} about it.
We make this notion precise and answer the following research question:
\textbf{(RQ1)} \textit{What constitutes structural evidence in $\Gi$ for a test triple $(h,r,t)$, and how does it translate to the relation graph $\Gri$ formulation?}

\paragraph{Half-link taxonomy.}
We decompose each test triple~$(h,r,t)$ into two \textbf{half-links}: the \textbf{query half}~$(h,r,?)$ and the \textbf{answer half}~$(?,r,t)$.
A half is \textbf{seen} when~$\Gi$ contains a \emph{witness} for it, a triple that attests it, and \textbf{unseen} otherwise; an unseen half is precisely where the model must generalize beyond the direct evidence in~$\Gi$.
Concretely, the query half~$(h,r,?)$ is seen iff $\exists\,e\in\Vi\setminus\{t\}:(h,r,e)\in\Ei$, and the answer half~$(?,r,t)$ is seen iff $\exists\,e\in\Vi\setminus\{h\}:(e,r,t)\in\Ei$.
Crossing the two binary outcomes yields a 2$\times$2 \textbf{half-link taxonomy}, namely \textbf{SQSA} (\textbf{S}een \textbf{Q}uery, \textbf{S}een \textbf{A}nswer), \textbf{SQUA} (\textbf{S}een \textbf{Q}uery, \textbf{U}nseen \textbf{A}nswer), \textbf{UQSA} (\textbf{U}nseen \textbf{Q}uery, \textbf{S}een \textbf{A}nswer), and \textbf{UQUA} (\textbf{U}nseen \textbf{Q}uery, \textbf{U}nseen   \textbf{A}nswer).
The taxonomy classifies each test triple $(h,r,t)\in\Eu$ relative to $\Gi$, independently of any particular method.

 \cref{fig:emnlp-quadrants} shows one running example: the inference graph $\Gi$ (solid) and the test graph $\Gu$ (dotted). Each scenario considers a single test triple (thick teal); a half is seen (thick orange) when $\Gi$ contains a witness triple for it.
(\textbf{SQSA}) The test triple $(\text{France},\,\emph{comprises},\,\text{Eiffel Tower})$ in \cref{fig:emnlp-quadrants}(a) relies on evidence already in $\Gi$: France is a
\emph{comprises}-source and Eiffel Tower a \emph{comprises}-target, each seen through an orange half-link.
(\textbf{SQUA}) The test triple $(\text{France},\,\emph{comprises},\,\text{Champ de Mars})$ in \cref{fig:emnlp-quadrants}(b) keeps a seen query-half, France still a \emph{comprises}-source,
but Champ de Mars is not a \emph{comprises}-target, so the answer half is unseen.
(\textbf{UQSA}) The dual of SQUA (\cref{fig:emnlp-quadrants}(c)): the answer half is seen, the query half unseen.
(\textbf{UQUA}) The test triple $(\text{Paris},\,\emph{comprises},\,\text{Champ de Mars})$ in \cref{fig:emnlp-quadrants}(d) has no such evidence: Paris is not a \emph{comprises}-source and
Champ de Mars not a \emph{comprises}-target, so neither half is seen and the model must generalize beyond the direct evidence in $\Gi$.
\cref{sec:proportions} measures how many test triples fall in each scenario across existing benchmarks.

  \paragraph{Relation-graph perspective.}
Since $\Gri$ is built exclusively from $\Gi$, it is the only structural input the relation encoder sees, so whether the four scenarios pose different challenges to the encoder depends on whether each leaves different structural evidence in $\Gri$. Unlike the taxonomy itself, how it maps onto $\Gri$ is method-dependent, since it hinges on the motif vocabulary of the method. We therefore ask, for each test triple $(h,r,t)\in\Eu$, which of the co-occurrence motifs it induces are already present in $\Gri$, and which might be missing.

 \cref{fig:relimpact} illustrates the two categorical poles on the running example. For \textbf{SQSA} (left), the test triple $(\text{France},\,\emph{comprises},\,\text{Eiffel Tower})$ would induce a \textit{t2t} edge between \emph{comprises} and \emph{hasLandmark}, sharing its tail Eiffel Tower with the \emph{hasLandmark} edge into it; this edge is already in $\Gri$, because by definition of SQSA Eiffel Tower has an incoming \emph{comprises} edge (from Champ de Mars).
 For \textbf{UQUA} (right), the test triple $(\text{Paris},\,\emph{comprises},\,\text{Champ de Mars})$ would induce a \textit{t2h} self-loop on \emph{comprises}, since its tail Champ de Mars is the head of an existing \emph{comprises} edge (out to Eiffel Tower); this edge is in $\Grt$, because by definition of UQUA neither half is seen, so no other \emph{comprises} edge ends at Champ de Mars or starts at Paris to close the chain. 
 In practice, though, it may still appear through incidental coverage, for example if $\Gi$ contained $(\text{Europe},\,\emph{comprises},\,\text{France})$, which would chain with the \emph{comprises} edge out of France to form the same \textit{t2h}, while leaving both halves of the test triple unseen. 
 We verify that this incidental coverage is common in ULTRA, and rare in TRIX (\cref{sec:relgraph}).
For \textbf{SQUA} and \textbf{UQSA}, 
one half is supplied directly, the other only incidentally.  
The breakdown for all four scenarios on the running example appears in~\cref{fig:relimpact-allq}.

\section{How many links from each scenario in current benchmarks?}\label{sec:proportions}
In this section, we analyze the scenario composition of these benchmarks and answer the following research question:
  \textbf{(RQ2)} \textit{How are test triples distributed across the four scenarios on the benchmarks used to evaluate KGFMs?} 
  
  \cref{tab:proportions} reports average scenario proportions across three benchmark families, ordered by how much of the inference graph $\Gi$ entity and relation name vocabularies already appears in the training graph $G$ the benchmark was built from:\footnote{Families 1, 2, and 3 are also known as \emph{transductive}, \emph{inductive entity}, and \emph{fully inductive} ~\citep{DBLP:conf/iclr/0001YM0Z24}, after the original setting each benchmark was created for; \cref{tab:app_datasets_transd,tab:app_datasets_inde,tab:app_datasets_indr} list the benchmarks in each.} ``\textit{Family~1}'' shares both entities and relation names, ``\textit{Family~2}'' shares relation names but introduces new entities, and ``\textit{Family~3}'' shares neither. We evaluate every benchmark in the inductive transfer setting of \cref{sec:found}; the family reflects how each benchmark was built, not what the model sees.
For each test triple $(h,r,t)\in\Eu$ scored in both directions, head prediction $(?,r,t)$ and tail prediction $(h,r,?)$ swap which half is the query and which is the answer; consequently, every test triple labelled \textbf{SQUA} for one direction is labelled \textbf{UQSA} for the other, and the two scenarios contain the same number of triples on a per-benchmark basis.
This identity is reflected in the Family~2 and Family~3 rows of \cref{tab:proportions}; the small gap in the Family~1 row stems from three of the sixteen benchmarks that score tail predictions only (\cref{sec:app-datasets}).

\begin{table}[!t]
\centering
\footnotesize
\setlength{\tabcolsep}{3pt}
\begin{adjustbox}{max width=\columnwidth}
\begin{tabular}{@{}lrrrr@{}}
\toprule
\textbf{Benchmarks ($N$)} & \textbf{SQSA} & \textbf{UQSA} & \textbf{SQUA} & \textbf{UQUA} \\
\midrule
Family 1 (16)            & $54.3\%$ & $21.7\%$ & $18.1\%$ & $5.9\%$  \\
Family 2 (18)            & $32.5\%$ & $28.8\%$ & $28.8\%$ & $9.9\%$  \\
Family 3 (23)            & $38.1\%$ & $28.0\%$ & $28.0\%$ & $5.9\%$  \\
\midrule
Overall (57)             & $40.9\%$ & $26.5\%$ & $25.5\%$ & $7.1\%$  \\
\bottomrule
\end{tabular}
\end{adjustbox}
\caption{\textbf{Fully unseen test triples (UQUA) account for under 10\% of test triples on average in every family}, while SQSA covers the largest share. $N$ = number of benchmarks averaged per row (unweighted). Per-benchmark breakdown in \cref{tab:app_quadrants}.}
\label{tab:proportions}
\end{table}

Across all three families, the largest share of test triples falls in \textbf{SQSA}, where both halves are seen, while fully unseen \textbf{UQUA} triples account for less than $10\%$ on average. See \cref{tab:app_quadrants} for per-benchmark figures.

\paragraph{An artefact of split design.}
The scenario proportions in \cref{tab:proportions} are a side effect of how these benchmarks were constructed, not a deliberate design target.
Per-benchmark SQSA proportions range from $0\%$ (NELL995~\citep{nell995}, Metafam~\citep{mtdea}) to $98\%$ (Hetionet~\citep{hetionet}), driven by heterogeneous split criteria: random train/test ratios for Family 1, entity disjointness for Family 2~\citep{DBLP:conf/icml/TeruDH20}, and relation-overlap fractions for Family 3~\citep{DBLP:conf/icml/LeeCW23}. 
Aggregate MRR mixes the scenarios in these split-driven proportions, whereas classifying each test triple with our taxonomy makes scenario-stratified MRR independent of that mix.

\begin{takeawaybox}[label={takeaway:proportions}]
\textbf{Takeaway 1.}
Aggregate MRR can hide where KGFMs suffer to generalize, due to different proportions of queries in SQSA, SQUA, UQSA, and UQUA.
We advise reporting it along our stratified MRR, to provide a finer-grained understanding.
\end{takeawaybox}

\section{What drives per-scenario KGFM performance?}\label{sec:arch}
A test triple decomposes into a query half and an answer half, each seen or unseen in $\Gi$ (\cref{sec:beyond}). The role each seen half plays in the link prediction performance of existing
KGFMs is unclear, so we answer the following research question:
\textbf{(RQ3)} \textit{How does each seen half-link signal, query or answer, affect KGFM link prediction performance?}
We aggregate scenario-stratified MRR over the 51 zero-shot KGs where every scenario is non-empty, excluding the usual pretraining graphs (FB15k237~\citep{DBLP:conf/acl-cvsc/ToutanovaC15}, CoDEx
Medium~\citep{codex}, WN18RR~\citep{wn18rr}) and the benchmarks with an empty scenario.

\paragraph{Architectural asymmetry of the two halves.}
Although the two halves are defined symmetrically over~$\Gi$, they play architecturally distinct roles in any GNN-based entity encoder. A \textbf{seen answer-half} provides an unambiguous positive signal: when $t$ has incoming $r$-typed edges in~$\Gi$, $r$-typed messages reach the representation of~$t$ through standard message passing regardless of the structure surrounding~$h$. A \textbf{seen query-half} is structurally ambiguous: when $h$ has outgoing $r$-typed edges in~$\Gi$, the representation of~$h$ encodes its existing $r$-tails as an implicit prior over plausible answers. This prior aligns with the true target~$t$ only if $t$ structurally resembles those existing $r$-tails, a condition the architecture does not guarantee~\citep{DBLP:conf/iclr/WuWZ0C22}; otherwise the prior competes against~$t$. The next two subsections confirm this asymmetry with a frozen baseline and a distractor diagnostic. We verify the asymmetry is not an artefact of node degree \citep{DBLP:conf/uai/MohamedPKA20} or of the relation-cardinality classes of \citet{DBLP:conf/nips/BordesUGWY13} in \cref{sec:app-degree,sec:app-cardinality}.

\paragraph{Architectural vs.\ learned origin.}
We isolate the architectural contribution with a \textbf{frozen baseline}, \ULTRArand: ULTRA with the relation encoder, the entity encoder, and the relation-graph initial features fixed at
random initialization, and only the score layer trained \citep{DBLP:journals/corr/abs-2203-02424, DBLP:journals/corr/abs-2502-00190}, so that any separation it shows across scenarios is attributable to the architecture rather than to learning.
We read each seen half off the scenario that isolates it: \textbf{UQSA}, where only the answer half is seen, and \textbf{SQUA}, where only the query half is seen.
On the 51-KG average (\cref{tab:uqsa_squa_ultra}) the frozen baseline already scores far higher on UQSA than on SQUA ($0.419$ vs.\ $0.129$): even without learning, a seen answer-half helps while a seen query-half does not.
Pre-training (\cref{sec:found})
lifts UQSA from $0.419$ to $0.547$ ($\Delta=+0.128$) while leaving SQUA essentially unchanged ($0.129 \to 0.133$).
Pre-training therefore amplifies the answer-half signal the architecture already supplies via the $r$-typed messages. 
Per-family rows and details on \ULTRArand\ implementation are in \cref{sec:app-ultra-ablation}.

\begin{table}[t]
\centering
\footnotesize
\setlength{\tabcolsep}{4pt}
\begin{adjustbox}{max width=\columnwidth}
\begin{tabular}{@{}ll ccc@{}}
\toprule
\textbf{Benchmarks} & \textbf{Scenario} & \textbf{ULTRA} & \textbf{\ULTRArand} & \textbf{$\Delta$} \\
\midrule
\multirow{2}{*}{Fam.\@ 1 (12)} & UQSA & 0.469 & 0.338 & \textbf{+0.131} \\
 & SQUA & 0.059 & 0.042 & +0.017 \\
\midrule
\multirow{2}{*}{Fam.\@ 2 (17)} & UQSA & 0.582 & 0.472 & \textbf{+0.110} \\
 & SQUA & 0.226 & 0.231 & $-$0.005 \\
\midrule
\multirow{2}{*}{Fam.\@ 3 (22)} & UQSA & 0.562 & 0.422 & \textbf{+0.140} \\
 & SQUA & 0.103 & 0.097 & +0.006 \\
\midrule
\multirow{2}{*}{All (51)} & UQSA & 0.547 & 0.419 & \textbf{+0.128} \\
 & SQUA & 0.133 & 0.129 & +0.004 \\
\bottomrule
\end{tabular}
\end{adjustbox}
\caption{\textbf{Pre-training mainly amplifies the seen answer signal}. 
UQSA scores far above SQUA, and pre-training gains are on UQSA. 
$\Delta$ = ULTRA $-$ \ULTRArand\ per scenario; bold marks the larger $\Delta$ per family.}
\label{tab:uqsa_squa_ultra}
\end{table}

\paragraph{Distractor diagnostic.}
A seen query-half   has existing $r$-answers in $\Gi$, the set $D(h,r)=\{e:(h,r,e)\in\Ei\}$. To test whether the seen-query signal acts as a distractor, we count how often a member of $D(h,r)$ outscores the held-out target $t$. Because the entities in $D(h,r)$ are themselves correct answers, the filtered protocol removes them from the ranking, so we read their scores from the raw output. 
A member of $D(h,r)$ outscores $t$ on most SQUA triples across \textit{Family 1}
benchmarks we selected (up to $91\%$ on WDsinger, $66\%$ on DBpedia100k): the seen-query prior ranks
the known answers of $h$ above the held-out target, competing against the correct prediction rather than supporting it. See \cref{sec:app-hown,tab:hown-rates} for more details.

\begin{takeawaybox}[label={takeaway:halflinks}]
\textbf{Takeaway 2.}
The two seen halves are asymmetric: the seen answer-half is a positive signal while the seen query-half is not.
Only scenario-stratified evaluation reveals whether a model overcomes it.
\end{takeawaybox}

\section{What drives UQUA generalization?}\label{sec:relgraph}
For a UQUA test triple, neither half-link is seen (\cref{fig:emnlp-quadrants}(d)), so the inference graph gives the entity encoder no direct signal (\cref{sec:arch}). The relation graph can still contribute: \cref{sec:beyond} showed that the motif a test triple would induce may already appear in $\Gri$ through incidental coverage, even when both halves are unseen.
Intuitively, a more expressive relation graph could raise this incidental coverage, and with it UQUA performance. We ask whether it always does:
\textbf{(RQ4)} \textit{Does a more expressive relation graph always improve generalization in UQUA, where $\Gi$ provides no direct half-link evidence?}
We compare the three KGFMs of \cref{sec:found} (ULTRA, MOTIF, TRIX) on the same 51 zero-shot KGs as \cref{sec:arch}, and additionally evaluate the single-pass ablation \TRIXnoiter~\citep{DBLP:conf/log/ZhangBGR24}. ULTRA, MOTIF, and \TRIXnoiter\ share the entity-level GNN \citep{DBLP:conf/nips/ZhuZXT21} and differ in how they construct the relation graph, so their comparison isolates the relation-graph design. 
Full TRIX adds iterative entity--relation refinement on top of the same relation graph of \TRIXnoiter, 
thus highlighting 
the contribution of the iterative mechanism rather than of per-entity granularity.

\paragraph{MOTIF leads UQUA among the three relation-graph designs.}
Among the three relation-graph designs, MOTIF posts the highest UQUA MRR in every benchmark family, while \TRIXnoiter\ falls below ULTRA (\cref{tab:uqua_main}). The pattern follows from whether the co-occurrence a UQUA test triple would induce is incidentally covered in $\Gri$. The higher-order motifs of MOTIF are entity-agnostic, so a co-occurrence $\Gi$ exposes incidentally covers any UQUA test triple that would induce it. \TRIXnoiter\ instead tags each co-occurrence with the entity that instantiates it in $\Gi$. A UQUA test triple instantiates its co-occurrence at a different entity, so the tagged motif does not match: the entity tag leaves the test triple uncovered, contributing no signal.
We confirm this with the incidental-coverage measurement introduced in \cref{sec:beyond}, computed per model and test triple (\cref{sec:app-relgraph-nadd}). Across four benchmarks spanning different UQUA shares and performances, the relation graph $\Gri$ of ULTRA covers $90.8\%$ to $99.5\%$ of induced $\Grt$ motifs, whereas the $\Gri$ of \TRIXnoiter\ covers under $0.6\%$ of induced $\Grt$ motifs (\cref{tab:relnadd-main}). The entity-tagged relation graph thus leaves \TRIXnoiter\ with little usable signal, which explains the UQUA result.

\paragraph{Iterative entity--relation coupling recovers UQUA.}
Full TRIX nevertheless closes this gap through a different route. Its iterative entity--relation coupling, learned during pre-training, likely aligns the representations of entities that participate in similar relations, supplying at zero-shot inference the entity-agnostic transfer that the entity-tagged relation graph alone cannot. The recovery is sharp, $+0.08$ MRR over \TRIXnoiter\ on the 51-KG average, far larger than the gain of iterative coupling on any other scenario (\cref{sec:app-iter-ablation}, \cref{tab:app_iterablation_perdataset}). Improving UQUA is therefore not only a matter of a more expressive relation graph: a pre-trained architectural mechanism such as iterative entity--relation coupling is a complementary lever.

\begin{takeawaybox}[label={takeaway:uqua}]
\textbf{Takeaway 3.}
Stratifying by scenario reveals that UQUA is where relation-graph design matters most. By looking at aggregate MRR alone, performance gains cannot be directly attributed to the relation-graph design.
\end{takeawaybox}

\begin{table}[t]
\centering
\footnotesize
\setlength{\tabcolsep}{3pt}
\begin{adjustbox}{max width=\columnwidth}
\begin{tabular}{@{}l cccc@{}}
\toprule
\textbf{Model} & \textbf{Fam. 1 (12)} & \textbf{Fam. 2 (17)} & \textbf{Fam. 3 (22)} & \textbf{All (51)} \\
\midrule
ULTRA & 0.202 & 0.316 & 0.214 & 0.245 \\
MOTIF & \textbf{0.216} & \textbf{0.352} & \textbf{0.229} & \textbf{0.267} \\
\TRIXnoiter & 0.145 & 0.256 & 0.159 & 0.188 \\
\bottomrule
\end{tabular}
\end{adjustbox}
\caption{\textbf{MOTIF has the relation-graph design that best transfers to unseen halves}, while \TRIXnoiter\ falls below ULTRA; full TRIX closes this gap through iterative entity--relation coupling (\cref{tab:app_iterablation_perdataset}). Unweighted mean UQUA MRR per benchmark family on the 51 zero-shot KGs; bold marks the best within a family.}
\label{tab:uqua_main}
\end{table}

\section{KGFM performance spreads widely}\label{sec:results}
We now evaluate the three KGFMs of \cref{sec:found} together under our scenario-stratified protocol, which we package as a reusable diagnostic,\footnote{We bring ULTRA, MOTIF, and TRIX into a single unified repository, together with the scenario-labelling code, so the protocol can be re-applied to new KGFMs and benchmarks; \url{https://github.com/cgregucci/KG-foundation-models}.} on the same 51 zero-shot KGs as in \cref{sec:arch}, using the publicly released pre-trained checkpoint of each model. We ask:
\textbf{(RQ5)} 
\textit{How do SoTA KGFMs perform zero-shot across the four scenarios 
and where does per-benchmark fine-tuning help?}

\begin{table*}[t]
\centering
\begin{minipage}{.6\textwidth}
    \small
\setlength{\tabcolsep}{6pt}
\begin{tabular}{@{}ll ccccc@{}}
\toprule
\textbf{Benchmarks} & \textbf{Model} & \textbf{Orig} & \textbf{SQSA} & \textbf{UQSA} & \textbf{SQUA} & \textbf{UQUA} \\
\midrule
\multirow{3}{*}{Fam.\@ 1 (12)}
 & ULTRA & \cellcolor{bgrey2!13}0.293 & \cellcolor{bgrey2!12}0.269 & \cellcolor{bgrey2!20}0.469 & \cellcolor{bgrey2!3}0.059 & \cellcolor{bgrey2!9}0.202 \\
 & MOTIF & \cellcolor{bgrey2!12}0.286 & \cellcolor{bgrey2!11}0.259 & \cellcolor{bgrey2!19}0.444 & \cellcolor{bgrey2!4}0.079 & \cellcolor{bgrey2!10}0.216 \\
 & TRIX  & \cellcolor{bgrey2!14}\textbf{0.314} & \cellcolor{bgrey2!12}\textbf{0.285} & \cellcolor{bgrey2!21}\textbf{0.491} & \cellcolor{bgrey2!4}\textbf{0.088} & \cellcolor{bgrey2!10}\textbf{0.225} \\
\midrule
\multirow{3}{*}{Fam.\@ 2 (17)}
 & ULTRA & \cellcolor{bgrey2!18}0.419 & \cellcolor{bgrey2!20}\textbf{0.487} & \cellcolor{bgrey2!24}0.582 & \cellcolor{bgrey2!10}0.226 & \cellcolor{bgrey2!14}0.316 \\
 & MOTIF & \cellcolor{bgrey2!18}0.422 & \cellcolor{bgrey2!20}0.473 & \cellcolor{bgrey2!24}0.569 & \cellcolor{bgrey2!11}0.254 & \cellcolor{bgrey2!15}\textbf{0.352} \\
 & TRIX  & \cellcolor{bgrey2!18}\textbf{0.435} & \cellcolor{bgrey2!20}0.483 & \cellcolor{bgrey2!24}\textbf{0.585} & \cellcolor{bgrey2!12}\textbf{0.273} & \cellcolor{bgrey2!15}0.343 \\
\midrule
\multirow{3}{*}{Fam.\@ 3 (22)}
 & ULTRA & \cellcolor{bgrey2!15}0.346 & \cellcolor{bgrey2!16}\textbf{0.385} & \cellcolor{bgrey2!23}0.562 & \cellcolor{bgrey2!5}0.103 & \cellcolor{bgrey2!10}0.214 \\
 & MOTIF & \cellcolor{bgrey2!15}0.349 & \cellcolor{bgrey2!16}0.372 & \cellcolor{bgrey2!23}0.545 & \cellcolor{bgrey2!7}0.142 & \cellcolor{bgrey2!10}0.229 \\
 & TRIX  & \cellcolor{bgrey2!16}\textbf{0.368} & \cellcolor{bgrey2!16}0.370 & \cellcolor{bgrey2!24}\textbf{0.579} & \cellcolor{bgrey2!8}\textbf{0.166} & \cellcolor{bgrey2!10}\textbf{0.235} \\
\midrule
\multirow{3}{*}{All (51)}
 & ULTRA & \cellcolor{bgrey2!15}0.358 & \cellcolor{bgrey2!17}\textbf{0.392} & \cellcolor{bgrey2!23}0.547 & \cellcolor{bgrey2!6}0.133 & \cellcolor{bgrey2!11}0.245 \\
 & MOTIF & \cellcolor{bgrey2!15}0.359 & \cellcolor{bgrey2!16}0.379 & \cellcolor{bgrey2!22}0.529 & \cellcolor{bgrey2!8}0.164 & \cellcolor{bgrey2!12}0.267 \\
 & TRIX  & \cellcolor{bgrey2!16}\textbf{0.378} & \cellcolor{bgrey2!16}0.388 & \cellcolor{bgrey2!23}\textbf{0.561} & \cellcolor{bgrey2!8}\textbf{0.184} & \cellcolor{bgrey2!12}\textbf{0.268} \\
\bottomrule
\end{tabular}
\end{minipage}\hfill\begin{minipage}{.35\textwidth}
\caption{  \textbf{UQSA is the highest-MRR scenario and SQUA the lowest for every model and family}, and the \textbf{Orig} aggregate sits closest to \textbf{SQSA}, the largest scenario, hiding the spread over the other three. Values are unweighted mean MRR per benchmark family on the zero-shot KGs with at least one triple in every scenario (per-family counts in parentheses); bold marks the best of the three models per family and scenario. Per-dataset MRR in \cref{sec:app-kgfm-mrr}.
    }
\label{tab:main_results}
\end{minipage}
\end{table*}

\paragraph{The per-scenario spread is as large as the aggregate.}
\cref{tab:main_results} reports scenario-stratified zero-shot MRR for ULTRA, MOTIF and TRIX. The ordering UQSA $>$ SQSA $>$ UQUA $>$ SQUA holds for every model and family. It follows from the asymmetry of the two halves (\cref{sec:arch}): a seen answer-half is an unambiguous positive signal, whereas a seen query-half is a distractor that competes against the target. UQSA carries the positive signal with no distractor and ranks highest; SQSA adds the distractor and ranks second; UQUA, with neither half seen, ranks third; and SQUA, where the distractor acts with no positive signal to offset it, ranks lowest, below even the all-unseen UQUA.

The spread this produces is wide: on the 51-KG average the strongest scenario exceeds the weakest by $0.37$ to $0.41$ MRR (UQSA vs.\ SQUA), as large as the \textbf{Orig} aggregate itself. The aggregate hides it, because \textbf{Orig} aggregate sits closest to \textbf{SQSA}, the largest scenario in current benchmarks (\cref{tab:proportions}).
This gap between the aggregate and the per-scenario view is clearest on UQUA, which isolates generalization beyond $\Gi$. ULTRA and MOTIF tie on the aggregate ($0.358$ vs.\ $0.359$) but not on UQUA, where MOTIF leads ($0.267$ vs.\ $0.245$); TRIX leads on the aggregate ($0.378$ vs.\ $0.359$) but its gain does not reach UQUA, where it ties MOTIF ($0.268$ vs.\ $0.267$), consistent with \cref{sec:relgraph}.

\paragraph{Fine-tuning reverses model ranking and partially recovers unseen answer-halves scenarios.}
The lower performance on the unseen answer-halves scenarios (SQUA, UQUA) has two possible sources: structure absent from $\Gi$, or structure present in $\Gi$ that the generic pre-trained weights do not extract. We disentangle them with per-benchmark fine-tuning on Family 3, the only family on which fine-tuning preserves the inductive transfer setting (\cref{sec:found}): there the training graph $G$ is disjoint from $\Gi$ in both entities and relation names.
Each pre-trained checkpoint is fine-tuned on the training graph $G$ of its benchmark following the fine-tuning protocol of ULTRA,\footnote{TRIX implements %
a zero-shot fallback which we do not enable so that any recovery is attributable to fine-tuning alone.} then message passing is performed on $\Gi$ exactly as in the zero-shot setting.
So comparing zero-shot and fine-tuned blocks in \cref{tab:finetune_ft_v2} isolates how much more in-$\Gi$ signal the fine-tuned weights extract (detailed results in \cref{sec:app-finetune}).
Since only the weights change, any unseen answer-half score that fine-tuning recovers is already extractable from $\Gi$.

On the seen answer-halves scenarios SQSA and UQSA, all models improve and the leader is unchanged, as
fine-tuning sharpens how each model exploits the seen-answer signal of \cref{sec:arch}. 
On the unseen answer-halves scenarios SQUA and UQUA the ranking instead flips: before, TRIX lead both, but after fine-tuning it falls to last on both, while the entity-agnostic ULTRA and MOTIF rise to the top: ULTRA goes from $0.214$ to $0.250$ on UQUA and MOTIF scores the best on SQUA.
Whether a model improves or regresses on the unseen answer-halves scenarios follows the relation-graph design of \cref{sec:relgraph}: fine-tuning sharpens the entity-agnostic structure of ULTRA and MOTIF, but degrades the per-entity structure of TRIX.

\begin{takeawaybox}[label={takeaway:aggregate}]
\textbf{Takeaway 4.}
Our half-link taxonomy exposes a wide spread in KGFM performance across scenarios. Fine-tuning further
shows that the unseen answer-half shortfall is partly an extraction gap, 
and thus a target for designing future KGFMs.
\end{takeawaybox}
\begin{table}[h]
\small
\centering
\setlength{\tabcolsep}{5pt}
\begin{adjustbox}{max width=\columnwidth}
\begin{tabular}{@{}cl cccc@{}}
\toprule
 & \textbf{Model} & \textbf{SQSA} & \textbf{UQSA} & \textbf{SQUA} & \textbf{UQUA} \\
\midrule
\multirow{3}{*}{\rotatebox{90}{\tiny\itshape Zero-shot}}
 & ULTRA & ${\color{petroil2}\mathbf{0.385}}$ & $0.562$                                    & $0.103$                                    & $0.214$                                    \\
 & MOTIF & $0.372$                            & $0.545$                                    & $0.142$                                    & $0.229$                                    \\
 & TRIX  & $0.370$                            & ${\color{gold2}\mathbf{0.579}}$           & ${\color{gold2}\mathbf{0.166}}$           & ${\color{gold2}\mathbf{0.235}}$           \\
\midrule
\multirow{3}{*}{\rotatebox{90}{\tiny\itshape Fine-tuned}}
 & ULTRA & ${\color{petroil2}\mathbf{0.399}}$ & $0.580$                                    & $0.149$                                    & ${\color{petroil2}\mathbf{0.250}}$        \\
 & MOTIF & $0.396$                            & $0.583$                                    & ${\color{lacamlightlilac}\mathbf{0.157}}$ & $0.235$                                    \\
 & TRIX  & $0.394$                            & ${\color{gold2}\mathbf{0.597}}$           & $0.115$                                    & $0.189$                                    \\
\bottomrule
\end{tabular}
\end{adjustbox}
\caption{\textbf{Fine-tuning flips the answer-unseen model ranking.} MRR on Family 3 benchmarks; in each block (zero-shot, fine-tuned) and scenario, the winning model is colored: \textcolor{petroil2}{\textbf{ULTRA}}, \textcolor{lacamlightlilac}{\textbf{MOTIF}}, \textcolor{gold2}{\textbf{TRIX}}.}
\label{tab:finetune_ft_v2}
\end{table}

\section{Conclusion}
Prior work measures KGFM generalization on the whole link, by aggregate MRR. We provide a finer-grained understanding by decomposing each link into two half-links, a query half and an answer half, each seen or unseen in the inference graph, into four scenarios. Stratified over them, performance varies by about $0.4$ MRR, a spread as wide as the aggregate yet invisible in it. The spread follows an asymmetry between the two halves, a seen answer-half being a positive signal and a seen query-half a distractor; the seen answer-half alone can be enough to predict a whole link. The taxonomy extends to the relation graph that GNN-based KGFMs build, where it acts as a proxy for when relation-graph expressiveness helps. Such expressiveness lifts the all-unseen scenario in particular, and only when the relation graph is entity-agnostic rather than per-entity.

The four scenarios occur in different, split-driven proportions, which the aggregate cannot disentangle. We therefore release our scenario-stratified protocol as a reusable diagnostic and recommend reporting it alongside the aggregate, since classifying each link by scenario makes this report independent of how benchmarks are split. Fine-tuning further locates the unseen answer-half shortfall as partly an extraction gap, signal already in the inference graph that current models leave unused, marking it as a concrete target for future KGFMs. Our analysis covers GNN-based KGFMs; extending it to non-GNN-based approaches such as FLOCK~\citep{DBLP:journals/corr/abs-2510-01510}, which replaces message passing with probabilistic random-walk ensembles, is a natural next step.

\section*{Limitations}

Our analysis covers GNN-based KGFMs, among which ULTRA, MOTIF, and TRIX. The asymmetry we identify between a seen answer-half and a seen query-half, and the finding that relation-graph expressiveness lifts the all-unseen scenario only when the relation graph is entity-agnostic, are therefore established for this model family. Yet on FLOCK, an architecture with neither message passing nor relation graph, preliminary results (\cref{sec:app-flock}) already reproduce the same pattern; a more extensive evaluation is needed but costly, at about 186$\times$ the inference compute of TRIX (\cref{tab:flock_fb10}). The phenomenon may therefore extend beyond the family our mechanism explains; understanding why is a separate line of work, as is whether text- and language-model-based link predictors exhibit it.

\section*{Author contributions}
CG conceived the initial idea that the inference graph supports test triples to different scenarios in SoTA KGFMs. CG, DH, and OT designed the half-link taxonomy, and OT performed the initial experiments to validate the idea. CG wrote the first draft of the manuscript, drew all the figures, designed and ran all the experiments, with the exception of FLOCK, ULTRA, and \ULTRArand, which were performed by OT.
SS suggested extending the analysis to the relation level, and AV had the intuition to characterize the taxonomy in terms of the relation graph. All authors critically revised the paper. AV and SS supervised all phases of the project and gave feedback.

\section*{Acknowledgements}
AV was supported by the ``UNREAL: Unified Reasoning Layer for Trustworthy ML'' project (EP/Y023838/1) selected by the ERC and funded by UKRI EPSRC.
CG and OT were funded by the CHIPS Joint Undertaking (JU) under grant agreement No. 101140087 (SMARTY), and by the German Federal Ministry of Education and Research (BMBF) under the sub-project with funding number 16MEE0444.
CG and OT acknowledge compute time on HoreKa HPC (NHR@KIT), funded by the BMBF and the MWK of Baden-Württemberg through the NHR program, with additional support from the DFG. CG worked on the paper partially during a research stay at the University of Edinburgh, funded by ELLIS Unit Stuttgart and by G-Research research grant CG20251209. DH was funded by the Deutsche Forschungsgemeinschaft (DFG, German Research Foundation) under Germany's
Excellence Strategy -- EXC 2120/2 -- 390831618. Halfling icon generate by Gemini.

\bibliography{lp}

\clearpage
\appendix
\numberwithin{table}{section}
\numberwithin{figure}{section}
\renewcommand{\topfraction}{0.9}
\renewcommand{\dbltopfraction}{0.9}
\renewcommand{\floatpagefraction}{0.85}
\renewcommand{\dblfloatpagefraction}{0.85}
\renewcommand{\textfraction}{0.08}
\setcounter{topnumber}{4}
\setcounter{dbltopnumber}{3}
\setcounter{totalnumber}{6}
\crefname{appendix}{App.}{Apps.}
\Crefname{appendix}{App.}{Apps.}
\let\oldsection\section
\renewcommand{\section}[1]{\oldsection{#1}\crefalias{section}{appendix}}
\let\oldsubsection\subsection
\renewcommand{\subsection}[1]{\oldsubsection{#1}\crefalias{subsection}{appendix}}

\section{Per-scenario relation-graph impact}
\label{sec:app-relimpact}
\begin{figure*}[t]
  \centering
  \includegraphics[width=\linewidth]{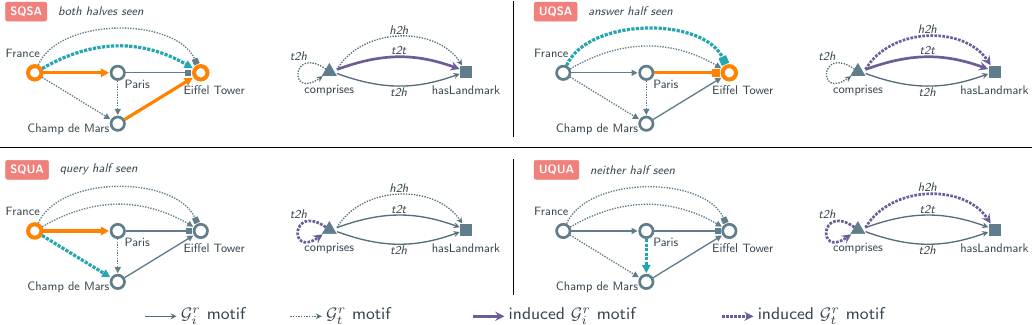}
\caption{\textbf{Only in SQSA are the motifs a test triple induces present in $\Gri$ by construction}, while in the other scenarios they appear only through incidental co-occurrences or are
missing. For each scenario we draw the four co-occurrence motifs between \emph{comprises} and \emph{hasLandmark}, solid if present in $\Gri$ and dotted if only in $\Grt$; those induced by
the defining test triple (teal, left graph of each pair) are thick \textcolor{lacamlilac}{\textbf{purple}}. In \textbf{SQSA} the induced \textit{t2t} is present by construction, since the
seen answer-half gives the answer entity an incoming \emph{comprises} edge. In \textbf{UQSA} the induced \textit{t2t} is still present, but only through an incidental \emph{comprises} edge
into the answer entity. In \textbf{SQUA} and \textbf{UQUA} the induced \textit{h2h} or \textit{t2h} self-loop is missing (purple dotted), recoverable only through incidental co-occurrences
(see \cref{sec:relgraph}).}
\label{fig:relimpact-allq}
\end{figure*}

The four scenarios leave different structural evidence in the relation
graph $\Gri$.
\Cref{fig:relimpact-allq} extends the SQSA/UQUA contrast of
\cref{fig:relimpact} to all four scenarios on the running example,
drawing for each the co-occurrence motifs the test triple induces and
whether they are already present in $\Gri$. In the two intermediate
scenarios, SQUA and UQSA, exactly one half is seen, so the seen half
supplies its motifs directly while the unseen half is covered only
incidentally.

\section{KGFM scenario-stratified MRR}
\label{sec:app-kgfm-mrr}
\paragraph{Scenario labelling and MRR aggregation.}
  Every test triple $(h,r,t)$ is scored in both directions and labelled
  independently in each. Tail prediction $(h,r,?)$ has query half $(h,r)$ and
  answer half $(r,t)$; head prediction $(?,r,t)$ is encoded with the inverse
  relation, giving query half $(t,\,r{+}|\mathcal{R}_i|)$ and answer half
  $(r{+}|\mathcal{R}_i|,\,h)$. A half is \emph{seen} when its (entity, relation)
  pair has a matching edge in the inference graph $\Gi$ (\cref{sec:beyond}), and
  the scenario of the directed query follows from whether its query and answer
  halves are seen. We report, per scenario, the mean reciprocal rank over all
  directed queries in the scenario; this pooling weights the head and tail
  directions by their counts, keeping each scenario consistent with the
  \textbf{Orig} aggregate, the original unstratified MRR pooled over all directed
  queries. We then average each per-scenario MRR unweighted over the benchmarks of
  a family. For SQSA, UQUA, and \textbf{Orig} the two directions are balanced
  ($n_{\text{head}}=n_{\text{tail}}$) and the weighting has no effect. SQUA and
  UQSA are the exception: the inverse swaps the two halves of a triple, so a
  triple that is SQUA under tail prediction is UQSA under head prediction; the two
  scenarios then cover the same triples (\cref{sec:proportions}) in opposite
  directions ($n_{\text{head}}\neq n_{\text{tail}}$), and are averaged
  according to their weight.

  \paragraph{Aggregate and per-dataset results.}
  \cref{tab:familyavg_results} reports the \textbf{Orig} MRR per benchmark family
  and pooled over all 57 benchmarks; on the full suite ULTRA and MOTIF are
  essentially tied ($0.368$ vs.\ $0.369$) and TRIX leads ($0.388$), consistent
  with the published numbers.
  \cref{tab:app_kgfm_perdataset} gives the full per-dataset scenario-stratified MRR
  for ULTRA, MOTIF, and TRIX.

\begin{table*}[t]
\centering
\small
\begin{tabular}{@{}lccc|cc|c@{}}
\toprule
\textbf{Model} & \textbf{Family 1 (13)} & \textbf{Family 2 (18)} & \textbf{Family 3 (23)} & \textbf{Total (54)} & \textbf{Pretrain (3)} & \textbf{All (57)} \\
\midrule
ULTRA & 0.305 & 0.438 & 0.346 & 0.367 & 0.388 & 0.368 \\
MOTIF & 0.302 & 0.436 & 0.349 & 0.367 & 0.409 & 0.369 \\
TRIX & \textbf{0.326} & \textbf{0.455} & \textbf{0.367} & \textbf{0.387} & \textbf{0.409} & \textbf{0.388} \\
\bottomrule
\end{tabular}
 \caption{\textbf{TRIX is the strongest GNN-based KGFM on aggregate}. Reproduced results of ULTRA, MOTIF, TRIX using their original checkpoints. Scenario-stratified evaluation in \cref{tab:main_results} reveals
where this gain concentrates. Unweighted mean Orig MRR per benchmark family. Family 1 (13) excludes the three pretraining graphs (FB15k237, CoDEx Medium, WN18RR); Pretrain (3) reports them separately. All (57) pools the full benchmark suite (the 54 non-pretraining benchmarks plus the 3 pretraining graphs)}
  \label{tab:familyavg_results}
\label{tab:familyavg_results}
\end{table*}

\begin{table*}[t]
\centering
\scriptsize
\setlength{\tabcolsep}{3pt}
\begin{adjustbox}{max width=\textwidth}
\begin{tabular}{@{}c l ccccc ccccc ccccc@{}}
\toprule
& \multirow{2}{*}{\textbf{Dataset}}
  & \multicolumn{5}{c}{\textbf{ULTRA}}
  & \multicolumn{5}{c}{\textbf{MOTIF}}
  & \multicolumn{5}{c}{\textbf{TRIX}} \\
\cmidrule(lr){3-7} \cmidrule(lr){8-12} \cmidrule(lr){13-17}
 & & Orig & SQSA & UQSA & SQUA & UQUA & Orig & SQSA & UQSA & SQUA & UQUA & Orig & SQSA & UQSA & SQUA & UQUA \\
\midrule
\multirow{17}{*}{\rotatebox[origin=c]{90}{\textbf{Family 1}}} & CoDEx Small & 0.460 & 0.464 & \textbf{0.758} & 0.142 & 0.307 & \textbf{0.474} & \textbf{0.472} & 0.754 & 0.224 & \textbf{0.347} & 0.473 & 0.469 & 0.756 & \textbf{0.236} & 0.337 \\
 & WDsinger & 0.386 & 0.369 & 0.495 & 0.163 & 0.609 & 0.397 & 0.370 & 0.486 & 0.193 & \textbf{0.638} & \textbf{0.398} & \textbf{0.373} & \textbf{0.506} & \textbf{0.202} & 0.592 \\
 & FB15k237\_10 & 0.240 & \textbf{0.129} & 0.470 & 0.011 & 0.033 & 0.236 & 0.127 & 0.458 & 0.012 & \textbf{0.046} & \textbf{0.246} & 0.127 & \textbf{0.481} & \textbf{0.014} & 0.043 \\
 & FB15k237\_20 & 0.268 & \textbf{0.160} & 0.544 & 0.014 & 0.074 & 0.259 & 0.152 & 0.523 & \textbf{0.018} & \textbf{0.088} & \textbf{0.269} & 0.156 & \textbf{0.548} & 0.017 & 0.081 \\
 & FB15k237\_50 & \textbf{0.322} & \textbf{0.221} & \textbf{0.666} & 0.031 & 0.177 & 0.312 & 0.213 & 0.644 & 0.034 & \textbf{0.185} & 0.321 & 0.220 & 0.663 & \textbf{0.036} & 0.179 \\
 & NELL23k & \textbf{0.245} & \textbf{0.197} & 0.532 & 0.037 & \textbf{0.242} & 0.220 & 0.175 & 0.492 & 0.043 & 0.164 & 0.237 & 0.180 & \textbf{0.535} & \textbf{0.049} & 0.188 \\
 & AristoV4 & 0.166 & 0.229 & 0.130 & 0.010 & 0.012 & 0.096 & 0.141 & 0.038 & 0.007 & 0.015 & \textbf{0.181} & \textbf{0.244} & \textbf{0.148} & \textbf{0.021} & \textbf{0.036} \\
 & Hetionet & 0.250 & 0.253 & 0.143 & 0.016 & 0.014 & 0.256 & 0.260 & 0.156 & 0.018 & 0.058 & \textbf{0.279} & \textbf{0.283} & \textbf{0.204} & \textbf{0.023} & \textbf{0.189} \\
 & NELL995 & 0.452 & -- & 0.753 & 0.106 & 0.526 & \textbf{0.491} & -- & 0.739 & 0.144 & \textbf{0.651} & 0.472 & -- & \textbf{0.799} & \textbf{0.162} & 0.444 \\
 & CoDEx Large & 0.328 & 0.314 & 0.622 & 0.065 & 0.338 & \textbf{0.339} & \textbf{0.318} & 0.624 & 0.100 & \textbf{0.362} & 0.335 & 0.310 & \textbf{0.625} & \textbf{0.104} & 0.351 \\
 & ConceptNet100k & 0.049 & 0.047 & 0.098 & 0.018 & \textbf{0.039} & 0.019 & 0.016 & 0.035 & 0.023 & 0.007 & \textbf{0.193} & \textbf{0.216} & \textbf{0.202} & \textbf{0.026} & 0.010 \\
 & DBpedia100k & 0.391 & 0.413 & 0.575 & 0.145 & 0.333 & 0.382 & 0.394 & 0.533 & 0.196 & 0.367 & \textbf{0.427} & \textbf{0.422} & \textbf{0.623} & \textbf{0.252} & \textbf{0.413} \\
 & YAGO310 & 0.412 & 0.432 & 0.595 & 0.053 & 0.241 & \textbf{0.441} & \textbf{0.465} & 0.590 & 0.075 & \textbf{0.321} & 0.409 & 0.425 & \textbf{0.600} & \textbf{0.079} & 0.277 \\
 & FB15k237$^{\dagger}$ & 0.354 & 0.338 & 0.702 & 0.088 & 0.293 & 0.346 & 0.328 & 0.677 & 0.101 & \textbf{0.307} & \textbf{0.362} & \textbf{0.339} & \textbf{0.704} & \textbf{0.128} & 0.306 \\
 & CoDEx Medium$^{\dagger}$ & 0.354 & 0.342 & \textbf{0.668} & 0.073 & 0.553 & \textbf{0.363} & \textbf{0.345} & 0.656 & 0.130 & \textbf{0.587} & 0.360 & 0.337 & 0.660 & \textbf{0.140} & 0.575 \\
 & WN18RR$^{\dagger}$ & 0.457 & 0.824 & 0.428 & 0.278 & 0.355 & \textbf{0.518} & \textbf{0.837} & \textbf{0.471} & 0.399 & \textbf{0.391} & 0.506 & 0.826 & 0.447 & \textbf{0.402} & 0.370 \\
\cmidrule(lr){2-17}
 & \textit{Average} & 0.305 & 0.269 & 0.491 & 0.062 & 0.227 & 0.302 & 0.259 & 0.467 & 0.084 & \textbf{0.250} & \textbf{0.326} & \textbf{0.285} & \textbf{0.515} & \textbf{0.094} & 0.242 \\
\midrule
\multirow{19}{*}{\rotatebox[origin=c]{90}{\textbf{Family 2}}} & FB v1 & 0.491 & 0.626 & 0.706 & 0.187 & 0.472 & 0.505 & 0.618 & 0.668 & \textbf{0.241} & \textbf{0.533} & \textbf{0.515} & \textbf{0.632} & \textbf{0.720} & 0.220 & 0.519 \\
 & FB v2 & 0.511 & \textbf{0.590} & 0.701 & 0.227 & 0.466 & 0.511 & 0.551 & 0.687 & 0.265 & \textbf{0.528} & \textbf{0.525} & 0.574 & \textbf{0.724} & \textbf{0.275} & 0.485 \\
 & FB v3 & 0.488 & 0.556 & 0.676 & 0.208 & 0.472 & 0.500 & 0.548 & 0.669 & \textbf{0.249} & \textbf{0.518} & \textbf{0.501} & \textbf{0.560} & \textbf{0.689} & 0.242 & 0.466 \\
 & FB v4 & 0.482 & 0.531 & 0.664 & 0.214 & 0.466 & 0.487 & 0.521 & 0.650 & \textbf{0.245} & \textbf{0.520} & \textbf{0.493} & \textbf{0.535} & \textbf{0.674} & 0.242 & 0.471 \\
 & WN v1 & 0.656 & 0.780 & 0.705 & 0.643 & 0.352 & 0.681 & \textbf{0.780} & \textbf{0.749} & 0.696 & 0.365 & \textbf{0.699} & 0.761 & 0.743 & \textbf{0.756} & \textbf{0.435} \\
 & WN v2 & 0.672 & \textbf{0.778} & \textbf{0.743} & 0.651 & 0.440 & 0.663 & 0.740 & 0.717 & 0.658 & 0.479 & \textbf{0.678} & 0.756 & 0.710 & \textbf{0.693} & \textbf{0.496} \\
 & WN v3 & 0.388 & \textbf{0.587} & 0.500 & 0.246 & 0.277 & \textbf{0.420} & 0.579 & \textbf{0.525} & 0.289 & 0.335 & 0.418 & 0.576 & 0.449 & \textbf{0.348} & \textbf{0.360} \\
 & WN v4 & 0.635 & \textbf{0.829} & 0.615 & 0.559 & 0.432 & 0.640 & 0.824 & \textbf{0.616} & 0.580 & 0.440 & \textbf{0.648} & 0.826 & 0.606 & \textbf{0.600} & \textbf{0.468} \\
 & NELL v1 & 0.756 & -- & \textbf{1.000} & 0.513 & -- & 0.669 & -- & 0.993 & 0.345 & -- & \textbf{0.806} & -- & 0.996 & \textbf{0.616} & -- \\
 & NELL v2 & 0.550 & 0.619 & \textbf{0.797} & 0.264 & 0.298 & 0.564 & \textbf{0.624} & 0.777 & 0.313 & \textbf{0.352} & \textbf{0.569} & 0.615 & 0.787 & \textbf{0.353} & 0.343 \\
 & NELL v3 & 0.538 & \textbf{0.577} & \textbf{0.855} & 0.218 & 0.311 & 0.533 & 0.540 & 0.820 & 0.270 & \textbf{0.393} & \textbf{0.558} & 0.566 & 0.840 & \textbf{0.318} & 0.361 \\
 & NELL v4 & 0.488 & 0.543 & 0.741 & 0.145 & 0.263 & 0.503 & 0.546 & 0.714 & 0.217 & \textbf{0.357} & \textbf{0.538} & \textbf{0.574} & \textbf{0.764} & \textbf{0.278} & 0.327 \\
 & ILPC Small & 0.298 & \textbf{0.326} & 0.503 & 0.042 & 0.316 & 0.296 & 0.319 & 0.498 & 0.051 & \textbf{0.325} & \textbf{0.303} & 0.315 & \textbf{0.522} & \textbf{0.062} & 0.318 \\
 & ILPC Large & 0.296 & \textbf{0.292} & 0.570 & 0.031 & 0.228 & 0.285 & 0.285 & 0.537 & 0.039 & \textbf{0.247} & \textbf{0.307} & 0.285 & \textbf{0.597} & \textbf{0.045} & 0.232 \\
 & HM 1k & \textbf{0.078} & \textbf{0.101} & \textbf{0.133} & 0.023 & 0.027 & 0.063 & 0.074 & 0.106 & 0.022 & \textbf{0.031} & 0.072 & 0.091 & 0.118 & \textbf{0.027} & 0.020 \\
 & HM 3k & 0.065 & \textbf{0.070} & 0.109 & \textbf{0.025} & 0.038 & 0.055 & 0.042 & 0.096 & 0.019 & \textbf{0.052} & \textbf{0.069} & 0.065 & \textbf{0.119} & 0.024 & 0.045 \\
 & HM 5k & 0.060 & \textbf{0.058} & 0.094 & \textbf{0.027} & \textbf{0.057} & 0.050 & 0.042 & 0.081 & 0.021 & 0.044 & \textbf{0.062} & 0.052 & \textbf{0.108} & 0.023 & 0.035 \\
 & IndigoBM & 0.428 & 0.418 & \textbf{0.783} & 0.126 & 0.464 & 0.426 & 0.416 & 0.764 & 0.135 & \textbf{0.473} & \textbf{0.436} & \textbf{0.426} & 0.783 & \textbf{0.144} & 0.449 \\
\cmidrule(lr){2-17}
 & \textit{Average} & 0.438 & \textbf{0.487} & 0.605 & 0.242 & 0.316 & 0.436 & 0.473 & 0.593 & 0.259 & \textbf{0.352} & \textbf{0.455} & 0.483 & \textbf{0.608} & \textbf{0.292} & 0.343 \\
\midrule
\multirow{24}{*}{\rotatebox[origin=c]{90}{\textbf{Family 3}}} & FB-25 & 0.387 & \textbf{0.386} & 0.608 & 0.135 & 0.444 & 0.384 & 0.370 & 0.587 & 0.151 & \textbf{0.501} & \textbf{0.393} & 0.384 & \textbf{0.614} & \textbf{0.155} & 0.466 \\
 & FB-50 & 0.334 & \textbf{0.355} & \textbf{0.501} & 0.100 & 0.376 & \textbf{0.338} & 0.348 & 0.473 & \textbf{0.126} & \textbf{0.428} & 0.334 & 0.341 & 0.497 & 0.119 & 0.390 \\
 & FB-75 & 0.397 & \textbf{0.444} & \textbf{0.625} & 0.111 & 0.289 & 0.399 & 0.431 & 0.609 & \textbf{0.134} & \textbf{0.360} & \textbf{0.401} & 0.437 & 0.622 & 0.130 & 0.334 \\
 & FB-100 & \textbf{0.446} & \textbf{0.481} & \textbf{0.671} & 0.132 & \textbf{0.317} & 0.428 & 0.462 & 0.639 & 0.141 & 0.254 & 0.436 & 0.465 & 0.660 & \textbf{0.144} & 0.288 \\
 & WK-25 & 0.310 & \textbf{0.346} & 0.490 & 0.102 & 0.166 & \textbf{0.311} & 0.334 & 0.491 & \textbf{0.114} & \textbf{0.206} & 0.305 & 0.319 & \textbf{0.499} & 0.114 & 0.189 \\
 & WK-50 & \textbf{0.175} & \textbf{0.184} & \textbf{0.274} & \textbf{0.059} & \textbf{0.182} & 0.163 & 0.169 & 0.259 & 0.057 & 0.163 & 0.166 & 0.167 & 0.268 & 0.059 & 0.175 \\
 & WK-75 & \textbf{0.386} & \textbf{0.472} & \textbf{0.597} & 0.069 & 0.134 & 0.366 & 0.438 & 0.576 & 0.073 & 0.133 & 0.368 & 0.435 & 0.580 & \textbf{0.081} & \textbf{0.137} \\
 & WK-100 & 0.178 & \textbf{0.263} & 0.242 & 0.009 & 0.019 & 0.164 & 0.238 & 0.219 & 0.018 & \textbf{0.027} & \textbf{0.188} & 0.261 & \textbf{0.269} & \textbf{0.022} & 0.027 \\
 & NL-0 & 0.364 & \textbf{0.431} & \textbf{0.656} & 0.132 & 0.158 & 0.324 & 0.356 & 0.610 & 0.119 & 0.125 & \textbf{0.385} & 0.385 & 0.650 & \textbf{0.236} & \textbf{0.168} \\
 & NL-25 & \textbf{0.399} & \textbf{0.390} & \textbf{0.769} & 0.115 & 0.097 & 0.348 & 0.359 & 0.680 & 0.083 & 0.071 & 0.377 & 0.317 & 0.724 & \textbf{0.132} & \textbf{0.123} \\
 & NL-50 & 0.394 & \textbf{0.437} & 0.757 & 0.086 & 0.081 & 0.373 & 0.387 & 0.739 & 0.083 & 0.073 & \textbf{0.404} & 0.357 & \textbf{0.829} & \textbf{0.104} & \textbf{0.091} \\
 & NL-75 & \textbf{0.355} & \textbf{0.394} & 0.626 & 0.126 & 0.137 & 0.314 & 0.341 & 0.555 & 0.121 & 0.116 & 0.351 & 0.322 & \textbf{0.653} & \textbf{0.155} & \textbf{0.165} \\
 & NL-100 & \textbf{0.469} & \textbf{0.511} & 0.767 & 0.140 & 0.214 & 0.438 & 0.445 & 0.723 & 0.158 & 0.139 & 0.468 & 0.451 & \textbf{0.773} & \textbf{0.191} & \textbf{0.261} \\
 & Metafam & 0.344 & -- & 0.178 & \textbf{0.510} & -- & \textbf{0.344} & -- & 0.416 & 0.272 & -- & 0.341 & -- & \textbf{0.446} & 0.236 & -- \\
 & FBNELL & \textbf{0.480} & \textbf{0.485} & \textbf{0.789} & 0.168 & 0.467 & 0.469 & 0.453 & 0.756 & 0.198 & \textbf{0.487} & 0.473 & 0.453 & 0.757 & \textbf{0.214} & 0.486 \\
 & Wiki MT1 tax & 0.240 & 0.281 & \textbf{0.486} & 0.004 & 0.003 & 0.325 & \textbf{0.304} & 0.462 & 0.217 & 0.003 & \textbf{0.358} & 0.251 & 0.463 & \textbf{0.300} & \textbf{0.015} \\
 & Wiki MT1 health & 0.297 & 0.446 & 0.480 & 0.055 & 0.301 & 0.326 & 0.458 & 0.521 & 0.081 & 0.301 & \textbf{0.376} & \textbf{0.480} & \textbf{0.627} & \textbf{0.086} & \textbf{0.301} \\
 & Wiki MT2 org & 0.084 & 0.079 & 0.221 & 0.035 & 0.002 & \textbf{0.092} & \textbf{0.079} & 0.279 & \textbf{0.098} & 0.002 & 0.091 & 0.074 & \textbf{0.349} & 0.091 & \textbf{0.003} \\
 & Wiki MT2 sci & 0.254 & 0.268 & \textbf{0.494} & 0.029 & 0.002 & 0.286 & \textbf{0.284} & 0.468 & 0.130 & 0.011 & \textbf{0.323} & 0.259 & 0.475 & \textbf{0.233} & \textbf{0.016} \\
 & Wiki MT3 art & 0.251 & 0.268 & 0.408 & 0.115 & 0.060 & 0.269 & \textbf{0.286} & 0.450 & 0.110 & 0.072 & \textbf{0.284} & 0.271 & \textbf{0.465} & \textbf{0.144} & \textbf{0.106} \\
 & Wiki MT3 infra & 0.596 & \textbf{0.662} & 0.812 & 0.277 & 0.601 & \textbf{0.658} & 0.645 & 0.812 & \textbf{0.531} & 0.618 & 0.655 & 0.651 & \textbf{0.817} & 0.501 & \textbf{0.637} \\
 & Wiki MT4 sci & \textbf{0.293} & \textbf{0.326} & 0.434 & 0.011 & 0.254 & 0.283 & 0.308 & 0.423 & 0.029 & \textbf{0.493} & 0.290 & 0.314 & \textbf{0.441} & \textbf{0.037} & 0.367 \\
 & Wiki MT4 health & 0.525 & 0.566 & 0.653 & 0.250 & 0.412 & 0.626 & 0.696 & 0.660 & 0.347 & \textbf{0.450} & \textbf{0.677} & \textbf{0.746} & \textbf{0.711} & \textbf{0.407} & 0.416 \\
\cmidrule(lr){2-17}
 & \textit{Average} & 0.346 & \textbf{0.385} & 0.545 & 0.121 & 0.214 & 0.349 & 0.372 & 0.540 & 0.147 & 0.229 & \textbf{0.367} & 0.370 & \textbf{0.574} & \textbf{0.169} & \textbf{0.235} \\
\bottomrule
\end{tabular}
\end{adjustbox}
  \caption{\textbf{Per-dataset scenario-stratified MRR for ULTRA, MOTIF, and TRIX.} All benchmark families in one table, separated by horizontal rules with a rotated family label per block;
  in each cell, bold marks the best of the three models. The italicised Average row per family is the unweighted mean over the zero-shot benchmarks of that family (pretraining rows excluded).
   $^{\dagger}$Used during KGFM pretraining; reported for completeness, not zero-shot. Empty scenarios (no test triples) are reported as ``--''.}
  \label{tab:app_kgfm_perdataset}
\end{table*}

\section{Relation-graph coverage of test triples}
\label{sec:app-relgraph-nadd}
 This section reports the incidental-coverage measurement that
  \cref{sec:relgraph} relies on, computed per model and test triple:
  how many of the co-occurrence motifs a test triple induces are already
  present in $\Gri$, and how many are missing.
  For each directed test triple $(h,r,t)$ we count the missing ones, the
  new relation-graph edges that appending this single triple to $\Gi$
  would induce at inference time.
  Let $R_0$ be the relation-graph edge set induced by the inference graph
  $\Gi$ (augmented with inverse relations) that the model conditions on
  at test time, and $R_1$ the same edge set after appending the test
  triple in both directions, forward $(h,r,t)$ and inverse
  $(t,\,r{+}|\mathcal{R}_i|,\,h)$; then $\texttt{nadded} = |R_1 \setminus R_0|$.
  For ULTRA, edges are $3$-tuples $(r_1, \tau, r_2)$ over the four binary
  motif types $\tau \in \{\hth,\tth,\ttt,\htt\}$; for TRIX, edges carry
  the witnessing entity and become $4$-tuples $(r_1, r_2, e, \tau)$ over
  the same four types \citep{DBLP:conf/log/ZhangBGR24}.
  Each test triple is scored in both directions and so appears as two
  directed triples; SQSA triples have $\texttt{nadded}\equiv 0$ by
  construction, since both halves of the triple are already seen in $\Gi$.

  \paragraph{Benchmark sub-selection.}
  We select four benchmarks that span two design axes: the share of UQUA
  in the test set, and the gap between the UQUA MRR of ULTRA and its
  aggregate (Orig) MRR. UQUA shares range from $1.7\%$ (FB15k237) through
  $2.8\%$ (ILPC2022SmallInductive) and $5.7\%$ (WKIngram:25) to $16.4\%$
  (WDsinger). The corresponding UQUA$-$Orig gap for ULTRA ranges from
  $-0.144$ (WKIngram:25, aggregate $0.310$) through $-0.061$ (FB15k237)
  and $+0.018$ (ILPC2022SmallInductive) up to $+0.223$ (WDsinger,
  aggregate $0.386$). Across the four the gap changes sign, placing UQUA
  below the aggregate on two benchmarks and above it on the other two.

  \paragraph{Per-scenario relation-graph coverage.}
  \cref{tab:relnadd-main} reports the diagnostic split by scenario for
  both ULTRA and TRIX side-by-side. We report, per (benchmark, split)
  cell: the number of directed test triples ($n_{\text{total}}$); the
  baseline edge count $|R_0|$ for each model (only on the Orig row, as
  $|R_0|$ is a property of the benchmark, not the split); the share of
  triples that induce at least one co-occurrence motif absent from $\Gri$
  ($\texttt{nadded}\geq 1$, a dotted $\Grt$ edge in \cref{fig:relimpact};
  $\%_{\geq 1}$); and the mean number of such motifs per triple.
  For ULTRA, almost every test triple already has all its induced motifs
  present in $\Gri$: its aggregate $\%_{\geq 1}$ (the Orig row) stays at
  or below $9.2\%$, and SQSA is $0\%$ by construction. For example, UQUA
  $\%_{\geq 1}$ peaks at $46.9\%$ on WKIngram:25, which is also the
  benchmark where UQUA falls furthest below the aggregate ($-0.144$). For
  TRIX, by contrast, whenever the witnessing entity of a motif differs
  between $\Gi$ and $\Gu$ that motif is absent from $\Gri$, so almost
  every test triple in SQUA, UQSA, and UQUA is missing at least one:
  $\%_{\geq 1}$ reaches exactly $100\%$ on the three benchmarks other than
  FB15k237 and $>99\%$ on FB15k237, with the mean number of missing
  motifs reaching $93$ per UQUA triple on FB15k237. Although the $|R_0|$
  of TRIX is $4\times$--$80\times$ larger than that of ULTRA (each ULTRA
  edge can split into many TRIX edges, one per witnessing entity), the
  entity-tagged construction makes the per-triple coverage strictly
  worse. This coverage gap is the structural explanation for the UQUA
  regression of \TRIXnoiter\ in \cref{tab:uqua_main}: without the
  iterative entity--relation updates that let representations transfer
  across the differing witnessing entities of $\Gi$ and $\Gu$, the
  relation graph that \TRIXnoiter\ relies on does not transfer to the
  UQUA scenario.

\begin{table*}[t]
\centering
\small
\setlength{\tabcolsep}{4pt}
\begin{tabular}{@{}l l r r r r r r r@{}}
\toprule
\multirow{2}{*}{\textbf{Benchmark}} & \multirow{2}{*}{\textbf{Split}}
  & \multirow{2}{*}{$n_{\text{total}}$}
  & \multicolumn{3}{c}{\textbf{ULTRA}}
  & \multicolumn{3}{c}{\textbf{TRIX}} \\
\cmidrule(lr){4-6} \cmidrule(lr){7-9}
 & & & $|R_0|$ & $\%_{\geq 1}$ & mean & $|R_0|$ & $\%_{\geq 1}$ & mean \\
\midrule
\multirow{5}{*}{WDsinger}
  & Orig &  4{,}406 & 21{,}920 &  6.40\% &  0.48 & 186{,}500 &  71.95\% &  8.76 \\
  & SQSA &  1{,}236 & \textendash &  0.00\% &  0.00 & \textendash &   0.00\% &  0.00 \\
  & SQUA &  1{,}224 & \textendash &  5.64\% &  0.32 & \textendash & 100.00\% &  9.03 \\
  & UQSA &  1{,}224 & \textendash &  5.64\% &  0.32 & \textendash & 100.00\% &  9.03 \\
  & UQUA & \phantom{0}\phantom{0}722 & \textendash & 19.94\% &  1.86 & \textendash & 100.00\% & 22.83 \\
\midrule
\multirow{5}{*}{ILPC2022SmallInductive}
  & Orig &  5{,}804 & \phantom{0}6{,}912 &  1.34\% &  0.06 & 198{,}938 &  55.44\% &  5.95 \\
  & SQSA &  2{,}586 & \textendash &  0.00\% &  0.00 & \textendash &   0.00\% &  0.00 \\
  & SQUA &  1{,}528 & \textendash &  1.77\% &  0.07 & \textendash & 100.00\% & 10.04 \\
  & UQSA &  1{,}528 & \textendash &  1.77\% &  0.07 & \textendash & 100.00\% & 10.04 \\
  & UQUA & \phantom{0}\phantom{0}162 & \textendash & 14.81\% &  0.74 & \textendash & 100.00\% & 23.70 \\
\midrule
\multirow{5}{*}{WKIngram:25}
  & Orig &  2{,}262 & \phantom{0}3{,}184 &  9.20\% &  0.46 & \phantom{0}15{,}148 &  56.68\% &  3.76 \\
  & SQSA & \phantom{0}\phantom{0}980 & \textendash &  0.00\% &  0.00 & \textendash &   0.00\% &  0.00 \\
  & SQUA & \phantom{0}\phantom{0}577 & \textendash & 12.82\% &  0.62 & \textendash & 100.00\% &  5.89 \\
  & UQSA & \phantom{0}\phantom{0}577 & \textendash & 12.82\% &  0.62 & \textendash & 100.00\% &  5.89 \\
  & UQUA & \phantom{0}\phantom{0}128 & \textendash & 46.88\% &  2.50 & \textendash & 100.00\% & 13.41 \\
\midrule
\multirow{5}{*}{FB15k237}
  & Orig & 40{,}932 & 108{,}240 &  0.53\% &  0.05 & 8{,}689{,}114 &  31.84\% & 17.96 \\
  & SQSA & 27{,}848 & \textendash &  0.00\% &  0.00 & \textendash &   0.00\% &  0.00 \\
  & SQUA &  6{,}203 & \textendash &  1.35\% &  0.14 & \textendash &  99.63\% & 54.14 \\
  & UQSA &  6{,}203 & \textendash &  1.35\% &  0.14 & \textendash &  99.63\% & 54.14 \\
  & UQUA & \phantom{0}\phantom{0}678 & \textendash &  7.08\% &  0.55 & \textendash &  99.41\% & 93.33 \\
\bottomrule
\end{tabular}
 \caption{\textbf{The relation graph of TRIX misses essentially every UQ$\ast$ and $\ast$UA triple.} ULTRA is missing a relation-graph edge for at most $9.2\%$ of test triples overall,
  whereas TRIX reaches $\%_{\geq 1}\geq 99.4\%$ across SQUA, UQSA, and UQUA and a mean of $13$--$93$ new edges per UQUA triple. $\texttt{nadded}=|R_1\setminus R_0|$ counts the relation-graph
  edges a single directed test triple induces that are missing from $R_0$; $\%_{\geq 1}$ is the share of triples with $\texttt{nadded}\geq 1$; mean is the average per-triple
  $\texttt{nadded}$. $|R_0|$ is shown once per benchmark on the Orig row. SQSA is $0$ by construction. The $|R_0|$ of TRIX is $4\times$--$80\times$ larger than that of ULTRA because each
  ULTRA edge $(r_1,\tau,r_2)$ splits into one TRIX edge per witnessing entity.}
  \label{tab:relnadd-main}
\end{table*}

\section{Iterative entity--relation coupling ablation (TRIX vs.\ \TRIXnoiter)}
\label{sec:app-iter-ablation}
\paragraph{The \TRIXnoiter\ variant.}
\TRIXnoiter\ is the single entity--relation pass ablation of TRIX from
\citet{DBLP:conf/log/ZhangBGR24}: it is identical to full TRIX, including the
same entity-tagged relation graph, except that the relation and entity networks
each run once in sequence rather than through the iterative entity--relation
refinement that full TRIX adds. We implement it and release the checkpoint. It
collapses on UQUA, where the entity that would witness a motif for the test
triple differs between $\Gi$ and $\Gu$ (\cref{sec:app-relgraph-nadd}), falling
below even ULTRA and MOTIF (\cref{tab:uqua_main}).

\paragraph{Per-dataset comparison.}
\cref{tab:app_iterablation_perdataset} gives the per-dataset comparison of TRIX
against \TRIXnoiter\ across all benchmark families, with per-family Average rows,
in the same unified format as the other per-dataset tables. The iterative
entity--relation updates let TRIX recover on UQUA, where \TRIXnoiter\ otherwise
falls behind.

\begin{table*}[t]
\centering
\scriptsize
\setlength{\tabcolsep}{3pt}
\begin{adjustbox}{max width=\textwidth}
\begin{tabular}{@{}c l ccccc ccccc@{}}
\toprule
& \multirow{2}{*}{\textbf{Dataset}}
  & \multicolumn{5}{c}{\textbf{TRIX}}
  & \multicolumn{5}{c}{\textbf{\TRIXnoiter}} \\
\cmidrule(lr){3-7} \cmidrule(lr){8-12}
 & & Orig & SQSA & UQSA & SQUA & UQUA & Orig & SQSA & UQSA & SQUA & UQUA \\
\midrule
\multirow{17}{*}{\rotatebox[origin=c]{90}{\textbf{Family 1}}} & CoDEx Small & \textbf{0.473} & \textbf{0.469} & \textbf{0.756} & 0.236 & \textbf{0.337} & 0.472 & 0.467 & 0.740 & \textbf{0.267} & 0.287 \\
 & WDsinger & \textbf{0.398} & \textbf{0.373} & \textbf{0.506} & \textbf{0.202} & \textbf{0.592} & 0.355 & 0.354 & 0.495 & 0.185 & 0.406 \\
 & FB15k237\_10 & \textbf{0.246} & \textbf{0.127} & \textbf{0.481} & 0.014 & \textbf{0.043} & 0.241 & 0.126 & 0.475 & \textbf{0.015} & 0.018 \\
 & FB15k237\_20 & \textbf{0.269} & 0.156 & \textbf{0.548} & 0.017 & \textbf{0.081} & 0.264 & \textbf{0.156} & 0.542 & \textbf{0.020} & 0.030 \\
 & FB15k237\_50 & \textbf{0.321} & 0.220 & 0.663 & 0.036 & \textbf{0.179} & 0.319 & \textbf{0.221} & \textbf{0.665} & \textbf{0.037} & 0.052 \\
 & NELL23k & \textbf{0.237} & \textbf{0.180} & \textbf{0.535} & \textbf{0.049} & \textbf{0.188} & 0.212 & 0.173 & 0.494 & 0.040 & 0.076 \\
 & AristoV4 & 0.181 & 0.244 & 0.148 & 0.021 & 0.036 & \textbf{0.198} & \textbf{0.258} & \textbf{0.183} & \textbf{0.032} & \textbf{0.044} \\
 & Hetionet & \textbf{0.279} & \textbf{0.283} & \textbf{0.204} & 0.023 & \textbf{0.189} & 0.234 & 0.236 & 0.182 & \textbf{0.025} & 0.038 \\
 & NELL995 & \textbf{0.472} & -- & \textbf{0.799} & \textbf{0.162} & \textbf{0.444} & 0.413 & -- & 0.793 & 0.139 & 0.243 \\
 & CoDEx Large & \textbf{0.335} & \textbf{0.310} & \textbf{0.625} & \textbf{0.104} & \textbf{0.351} & 0.328 & 0.303 & 0.615 & 0.099 & 0.323 \\
 & ConceptNet100k & \textbf{0.193} & \textbf{0.216} & \textbf{0.202} & 0.026 & 0.010 & 0.162 & 0.177 & 0.193 & \textbf{0.028} & \textbf{0.015} \\
 & DBpedia100k & \textbf{0.427} & \textbf{0.422} & \textbf{0.623} & \textbf{0.252} & \textbf{0.413} & 0.358 & 0.357 & 0.571 & 0.188 & 0.231 \\
 & YAGO310 & \textbf{0.409} & \textbf{0.425} & \textbf{0.600} & \textbf{0.079} & \textbf{0.277} & 0.273 & 0.260 & 0.600 & 0.077 & 0.215 \\
 & FB15k237$^{\dagger}$ & \textbf{0.362} & \textbf{0.339} & \textbf{0.704} & 0.128 & \textbf{0.306} & 0.358 & 0.337 & 0.699 & \textbf{0.130} & 0.177 \\
 & CoDEx Medium$^{\dagger}$ & \textbf{0.360} & \textbf{0.337} & \textbf{0.660} & \textbf{0.140} & \textbf{0.575} & 0.357 & 0.335 & 0.653 & 0.138 & 0.551 \\
 & WN18RR$^{\dagger}$ & 0.506 & 0.826 & 0.447 & 0.402 & 0.370 & \textbf{0.512} & \textbf{0.832} & \textbf{0.452} & \textbf{0.409} & \textbf{0.377} \\
\cmidrule(lr){2-12}
 & \textit{Average} & \textbf{0.326} & \textbf{0.285} & \textbf{0.515} & \textbf{0.094} & \textbf{0.242} & 0.295 & 0.257 & 0.503 & 0.089 & 0.152 \\
\midrule
\multirow{19}{*}{\rotatebox[origin=c]{90}{\textbf{Family 2}}} & FB v1 & \textbf{0.515} & \textbf{0.632} & \textbf{0.720} & 0.220 & \textbf{0.519} & 0.470 & 0.599 & 0.697 & \textbf{0.222} & 0.345 \\
 & FB v2 & \textbf{0.525} & \textbf{0.574} & \textbf{0.724} & \textbf{0.275} & \textbf{0.485} & 0.498 & 0.566 & 0.713 & 0.268 & 0.343 \\
 & FB v3 & \textbf{0.501} & \textbf{0.560} & \textbf{0.689} & \textbf{0.242} & \textbf{0.466} & 0.467 & 0.547 & 0.672 & 0.237 & 0.292 \\
 & FB v4 & \textbf{0.493} & \textbf{0.535} & \textbf{0.674} & \textbf{0.242} & \textbf{0.471} & 0.466 & 0.525 & 0.668 & 0.222 & 0.311 \\
 & WN v1 & \textbf{0.699} & 0.761 & \textbf{0.743} & \textbf{0.756} & \textbf{0.435} & 0.693 & \textbf{0.795} & 0.729 & 0.710 & 0.412 \\
 & WN v2 & \textbf{0.678} & 0.756 & 0.710 & \textbf{0.693} & \textbf{0.496} & 0.666 & \textbf{0.776} & \textbf{0.718} & 0.672 & 0.422 \\
 & WN v3 & \textbf{0.418} & \textbf{0.576} & \textbf{0.449} & \textbf{0.348} & \textbf{0.360} & 0.353 & 0.467 & 0.407 & 0.305 & 0.246 \\
 & WN v4 & 0.648 & 0.826 & 0.606 & \textbf{0.600} & \textbf{0.468} & \textbf{0.649} & \textbf{0.843} & \textbf{0.612} & 0.586 & 0.450 \\
 & NELL v1 & 0.806 & -- & 0.996 & 0.616 & -- & \textbf{0.810} & -- & \textbf{1.000} & \textbf{0.619} & -- \\
 & NELL v2 & \textbf{0.569} & \textbf{0.615} & \textbf{0.787} & \textbf{0.353} & \textbf{0.343} & 0.513 & 0.587 & 0.747 & 0.252 & 0.208 \\
 & NELL v3 & \textbf{0.558} & \textbf{0.566} & \textbf{0.840} & \textbf{0.318} & \textbf{0.361} & 0.530 & 0.563 & 0.792 & 0.292 & 0.241 \\
 & NELL v4 & \textbf{0.538} & 0.574 & \textbf{0.764} & \textbf{0.278} & \textbf{0.327} & 0.515 & \textbf{0.574} & 0.733 & 0.228 & 0.188 \\
 & ILPC Small & \textbf{0.303} & 0.315 & \textbf{0.522} & \textbf{0.062} & \textbf{0.318} & 0.300 & \textbf{0.315} & 0.515 & 0.059 & 0.291 \\
 & ILPC Large & \textbf{0.307} & \textbf{0.285} & \textbf{0.597} & \textbf{0.045} & \textbf{0.232} & 0.296 & 0.283 & 0.574 & 0.039 & 0.220 \\
 & HM 1k & 0.072 & 0.091 & 0.118 & \textbf{0.027} & \textbf{0.020} & \textbf{0.075} & \textbf{0.113} & \textbf{0.122} & 0.022 & 0.010 \\
 & HM 3k & 0.069 & 0.065 & 0.119 & 0.024 & \textbf{0.045} & \textbf{0.072} & \textbf{0.073} & \textbf{0.127} & \textbf{0.025} & 0.015 \\
 & HM 5k & 0.062 & 0.052 & 0.108 & \textbf{0.023} & \textbf{0.035} & \textbf{0.066} & \textbf{0.065} & \textbf{0.115} & 0.023 & 0.023 \\
 & IndigoBM & \textbf{0.436} & \textbf{0.426} & 0.783 & \textbf{0.144} & \textbf{0.449} & 0.412 & 0.394 & \textbf{0.784} & 0.144 & 0.339 \\
\cmidrule(lr){2-12}
 & \textit{Average} & \textbf{0.455} & \textbf{0.483} & \textbf{0.608} & \textbf{0.292} & \textbf{0.343} & 0.436 & 0.476 & 0.596 & 0.274 & 0.256 \\
\midrule
\multirow{24}{*}{\rotatebox[origin=c]{90}{\textbf{Family 3}}} & FB-25 & \textbf{0.393} & \textbf{0.384} & \textbf{0.614} & 0.155 & \textbf{0.466} & 0.359 & 0.380 & 0.603 & \textbf{0.157} & 0.179 \\
 & FB-50 & \textbf{0.334} & \textbf{0.341} & \textbf{0.497} & \textbf{0.119} & \textbf{0.390} & 0.288 & 0.341 & 0.479 & 0.116 & 0.116 \\
 & FB-75 & \textbf{0.401} & \textbf{0.437} & \textbf{0.622} & \textbf{0.130} & \textbf{0.334} & 0.369 & 0.425 & 0.613 & 0.120 & 0.108 \\
 & FB-100 & \textbf{0.436} & 0.465 & \textbf{0.660} & \textbf{0.144} & \textbf{0.288} & 0.425 & \textbf{0.467} & 0.647 & 0.125 & 0.137 \\
 & WK-25 & \textbf{0.305} & 0.319 & \textbf{0.499} & \textbf{0.114} & \textbf{0.189} & 0.293 & \textbf{0.322} & 0.485 & 0.082 & 0.161 \\
 & WK-50 & \textbf{0.166} & \textbf{0.167} & \textbf{0.268} & \textbf{0.059} & \textbf{0.175} & 0.157 & 0.162 & 0.257 & 0.053 & 0.140 \\
 & WK-75 & 0.368 & \textbf{0.435} & 0.580 & 0.081 & \textbf{0.137} & \textbf{0.369} & 0.425 & \textbf{0.597} & \textbf{0.096} & 0.088 \\
 & WK-100 & \textbf{0.188} & \textbf{0.261} & \textbf{0.269} & \textbf{0.022} & \textbf{0.027} & 0.183 & 0.257 & 0.260 & 0.018 & 0.013 \\
 & NL-0 & \textbf{0.385} & 0.385 & \textbf{0.650} & \textbf{0.236} & \textbf{0.168} & 0.354 & \textbf{0.391} & 0.644 & 0.174 & 0.096 \\
 & NL-25 & 0.377 & 0.317 & 0.724 & \textbf{0.132} & 0.123 & \textbf{0.392} & \textbf{0.352} & \textbf{0.768} & 0.109 & \textbf{0.127} \\
 & NL-50 & \textbf{0.404} & 0.357 & \textbf{0.829} & \textbf{0.104} & \textbf{0.091} & 0.400 & \textbf{0.390} & 0.809 & 0.094 & 0.064 \\
 & NL-75 & \textbf{0.351} & 0.322 & \textbf{0.653} & \textbf{0.155} & \textbf{0.165} & 0.326 & \textbf{0.344} & 0.579 & 0.138 & 0.111 \\
 & NL-100 & \textbf{0.468} & 0.451 & \textbf{0.773} & \textbf{0.191} & \textbf{0.261} & 0.437 & \textbf{0.481} & 0.726 & 0.113 & 0.202 \\
 & Metafam & 0.341 & -- & 0.446 & 0.236 & -- & \textbf{0.418} & -- & \textbf{0.560} & \textbf{0.276} & -- \\
 & FBNELL & \textbf{0.473} & \textbf{0.453} & \textbf{0.757} & \textbf{0.214} & \textbf{0.486} & 0.446 & 0.452 & 0.735 & 0.181 & 0.334 \\
 & Wiki MT1 tax & \textbf{0.358} & \textbf{0.251} & \textbf{0.463} & \textbf{0.300} & 0.015 & 0.300 & 0.248 & 0.456 & 0.177 & \textbf{0.020} \\
 & Wiki MT1 health & \textbf{0.376} & 0.480 & \textbf{0.627} & \textbf{0.086} & 0.301 & 0.371 & \textbf{0.490} & 0.620 & 0.076 & \textbf{0.302} \\
 & Wiki MT2 org & \textbf{0.091} & \textbf{0.074} & 0.349 & 0.091 & \textbf{0.003} & 0.088 & 0.070 & \textbf{0.350} & \textbf{0.104} & 0.002 \\
 & Wiki MT2 sci & \textbf{0.323} & \textbf{0.259} & \textbf{0.475} & \textbf{0.233} & \textbf{0.016} & 0.298 & 0.223 & 0.452 & 0.211 & 0.005 \\
 & Wiki MT3 art & 0.284 & 0.271 & \textbf{0.465} & \textbf{0.144} & 0.106 & \textbf{0.285} & \textbf{0.282} & 0.462 & 0.140 & \textbf{0.114} \\
 & Wiki MT3 infra & \textbf{0.655} & \textbf{0.651} & \textbf{0.817} & \textbf{0.501} & 0.637 & 0.636 & 0.628 & 0.806 & 0.478 & \textbf{0.645} \\
 & Wiki MT4 sci & \textbf{0.290} & \textbf{0.314} & \textbf{0.441} & \textbf{0.037} & \textbf{0.367} & 0.268 & 0.291 & 0.408 & 0.029 & 0.261 \\
 & Wiki MT4 health & \textbf{0.677} & \textbf{0.746} & \textbf{0.711} & \textbf{0.407} & \textbf{0.416} & 0.634 & 0.721 & 0.633 & 0.343 & 0.275 \\
\cmidrule(lr){2-12}
 & \textit{Average} & \textbf{0.367} & \textbf{0.370} & \textbf{0.574} & \textbf{0.169} & \textbf{0.235} & 0.352 & 0.370 & 0.563 & 0.148 & 0.159 \\
\bottomrule
\end{tabular}
\end{adjustbox}
\caption{\textbf{Per-dataset scenario-stratified MRR for TRIX and \TRIXnoiter\ (TRIX without the iterative entity--relation coupling).} All benchmark families in one table, separated by
horizontal rules with a rotated family label per block; in each cell, bold marks the better of the two models. The italicised Average row per family is the unweighted mean over the
zero-shot benchmarks of that family (pretraining rows excluded). $^{\dagger}$Used during KGFM pretraining; reported for completeness, not zero-shot. Empty scenarios (no test triples) are
reported as ``--''.}
\label{tab:app_iterablation_perdataset}
\end{table*}

\clearpage
\section{Frozen-baseline ablation\\(ULTRA vs.\ \ULTRArand)}
\label{sec:app-ultra-ablation}

\paragraph{The \ULTRArand\ baseline.}
\ULTRArand\ is our frozen-backbone ablation of ULTRA: we keep the entire backbone
at random initialization, the two NBFNets, every relational-convolution layer,
all layer-norm parameters, and the relation embeddings, and train only the
entity-side score head, a two-layer MLP ($128{\to}128{\to}1$) with $16{,}641$
parameters, $9.86\%$ of the $168{,}705$-parameter model; the remaining $90\%$
stay random. We pre-train this head for ten epochs on the
FB15k237\,+\,WN18RR\,+\,CoDEx~Medium mixture, following the exact same training
protocol as ULTRA, and release the checkpoint. It isolates how much of the
per-scenario behavior is fixed by the $r$-typed message-passing architecture
before any backbone weights are learned.

\paragraph{Per-dataset comparison.}
\cref{tab:app_ultraablation_perdataset} reports the per-dataset
scenario-stratified MRR for ULTRA against \ULTRArand\ across all benchmark
families. Even with the backbone frozen at random initialization, the
seen-answer advantage of \cref{sec:arch} is already visible under \ULTRArand\ on
average, indicating that it originates in the $r$-typed message passing of the
architecture rather than in learned weights.

\begin{table*}[t]
\centering
\scriptsize
\setlength{\tabcolsep}{3pt}
\begin{adjustbox}{max width=\textwidth}
\begin{tabular}{@{}c l ccccc ccccc@{}}
\toprule
& \multirow{2}{*}{\textbf{Dataset}}
  & \multicolumn{5}{c}{\textbf{ULTRA}}
  & \multicolumn{5}{c}{\textbf{\ULTRArand}} \\
\cmidrule(lr){3-7} \cmidrule(lr){8-12}
 & & Orig & SQSA & UQSA & SQUA & UQUA & Orig & SQSA & UQSA & SQUA & UQUA \\
\midrule
\multirow{17}{*}{\rotatebox[origin=c]{90}{\textbf{Family 1}}} & CoDEx Small & \textbf{0.460} & \textbf{0.464} & \textbf{0.758} & \textbf{0.142} & 0.307 & 0.355 & 0.362 & 0.557 & 0.086 & \textbf{0.390} \\
 & WDsinger & \textbf{0.386} & \textbf{0.369} & \textbf{0.495} & \textbf{0.163} & \textbf{0.609} & 0.294 & 0.266 & 0.335 & 0.114 & 0.577 \\
 & FB15k237\_10 & \textbf{0.240} & \textbf{0.129} & \textbf{0.470} & \textbf{0.011} & 0.033 & 0.215 & 0.096 & 0.431 & 0.009 & \textbf{0.047} \\
 & FB15k237\_20 & \textbf{0.268} & \textbf{0.160} & \textbf{0.544} & \textbf{0.014} & 0.074 & 0.225 & 0.113 & 0.477 & 0.011 & \textbf{0.083} \\
 & FB15k237\_50 & \textbf{0.322} & \textbf{0.221} & \textbf{0.666} & \textbf{0.031} & \textbf{0.177} & 0.239 & 0.150 & 0.526 & 0.017 & 0.164 \\
 & NELL23k & \textbf{0.245} & \textbf{0.197} & \textbf{0.532} & 0.037 & \textbf{0.242} & 0.171 & 0.153 & 0.343 & \textbf{0.040} & 0.141 \\
 & AristoV4 & \textbf{0.166} & \textbf{0.229} & \textbf{0.130} & \textbf{0.010} & 0.012 & 0.062 & 0.083 & 0.049 & 0.008 & \textbf{0.020} \\
 & Hetionet & \textbf{0.250} & \textbf{0.253} & \textbf{0.143} & \textbf{0.016} & 0.014 & 0.124 & 0.125 & 0.091 & 0.006 & \textbf{0.034} \\
 & NELL995 & \textbf{0.452} & -- & \textbf{0.753} & \textbf{0.106} & \textbf{0.526} & 0.164 & -- & 0.361 & 0.037 & 0.051 \\
 & CoDEx Large & \textbf{0.328} & \textbf{0.314} & \textbf{0.622} & \textbf{0.065} & \textbf{0.338} & 0.179 & 0.158 & 0.358 & 0.037 & 0.330 \\
 & ConceptNet100k & 0.049 & 0.047 & 0.098 & 0.018 & \textbf{0.039} & \textbf{0.075} & \textbf{0.069} & \textbf{0.166} & \textbf{0.037} & 0.018 \\
 & DBpedia100k & \textbf{0.391} & \textbf{0.413} & \textbf{0.575} & \textbf{0.145} & \textbf{0.333} & 0.196 & 0.191 & 0.279 & 0.105 & 0.289 \\
 & YAGO310 & \textbf{0.412} & \textbf{0.432} & \textbf{0.595} & \textbf{0.053} & 0.241 & 0.379 & 0.409 & 0.448 & 0.031 & \textbf{0.274} \\
 & FB15k237$^{\dagger}$ & \textbf{0.354} & \textbf{0.338} & \textbf{0.702} & \textbf{0.088} & \textbf{0.293} & 0.227 & 0.207 & 0.508 & 0.029 & 0.261 \\
 & CoDEx Medium$^{\dagger}$ & \textbf{0.354} & \textbf{0.342} & \textbf{0.668} & \textbf{0.073} & \textbf{0.553} & 0.229 & 0.222 & 0.415 & 0.053 & 0.547 \\
 & WN18RR$^{\dagger}$ & \textbf{0.457} & \textbf{0.824} & \textbf{0.428} & \textbf{0.278} & \textbf{0.355} & 0.391 & 0.771 & 0.303 & 0.246 & 0.326 \\
\cmidrule(lr){2-12}
 & \textit{Average} & \textbf{0.305} & \textbf{0.269} & \textbf{0.491} & \textbf{0.062} & \textbf{0.227} & 0.206 & 0.181 & 0.340 & 0.041 & 0.186 \\
\midrule
\multirow{19}{*}{\rotatebox[origin=c]{90}{\textbf{Family 2}}} & FB v1 & 0.491 & \textbf{0.626} & \textbf{0.706} & 0.187 & 0.472 & \textbf{0.500} & 0.550 & 0.646 & \textbf{0.232} & \textbf{0.643} \\
 & FB v2 & \textbf{0.511} & \textbf{0.590} & \textbf{0.701} & 0.227 & 0.466 & 0.490 & 0.513 & 0.635 & \textbf{0.262} & \textbf{0.576} \\
 & FB v3 & \textbf{0.488} & \textbf{0.556} & \textbf{0.676} & 0.208 & 0.472 & 0.448 & 0.485 & 0.581 & \textbf{0.218} & \textbf{0.530} \\
 & FB v4 & \textbf{0.482} & \textbf{0.531} & \textbf{0.664} & 0.214 & 0.466 & 0.446 & 0.468 & 0.589 & \textbf{0.225} & \textbf{0.530} \\
 & WN v1 & 0.656 & \textbf{0.780} & 0.705 & 0.643 & 0.352 & \textbf{0.674} & 0.749 & \textbf{0.722} & \textbf{0.721} & \textbf{0.393} \\
 & WN v2 & \textbf{0.672} & \textbf{0.778} & \textbf{0.743} & 0.651 & 0.440 & 0.654 & 0.745 & 0.690 & \textbf{0.669} & \textbf{0.447} \\
 & WN v3 & \textbf{0.388} & \textbf{0.587} & \textbf{0.500} & 0.246 & 0.277 & 0.349 & 0.522 & 0.384 & \textbf{0.259} & \textbf{0.304} \\
 & WN v4 & \textbf{0.635} & \textbf{0.829} & \textbf{0.615} & 0.559 & \textbf{0.432} & 0.615 & 0.794 & 0.576 & \textbf{0.568} & 0.425 \\
 & NELL v1 & 0.756 & -- & \textbf{1.000} & 0.513 & -- & \textbf{0.780} & -- & \textbf{1.000} & \textbf{0.560} & -- \\
 & NELL v2 & \textbf{0.550} & \textbf{0.619} & \textbf{0.797} & \textbf{0.264} & 0.298 & 0.457 & 0.505 & 0.623 & 0.258 & \textbf{0.299} \\
 & NELL v3 & \textbf{0.538} & \textbf{0.577} & \textbf{0.855} & \textbf{0.218} & \textbf{0.311} & 0.426 & 0.439 & 0.679 & 0.198 & 0.263 \\
 & NELL v4 & \textbf{0.488} & \textbf{0.543} & \textbf{0.741} & \textbf{0.145} & \textbf{0.263} & 0.400 & 0.448 & 0.570 & 0.145 & 0.231 \\
 & ILPC Small & \textbf{0.298} & \textbf{0.326} & \textbf{0.503} & \textbf{0.042} & \textbf{0.316} & 0.172 & 0.186 & 0.280 & 0.027 & 0.308 \\
 & ILPC Large & \textbf{0.296} & \textbf{0.292} & \textbf{0.570} & \textbf{0.031} & 0.228 & 0.172 & 0.157 & 0.336 & 0.018 & \textbf{0.228} \\
 & HM 1k & \textbf{0.078} & \textbf{0.101} & \textbf{0.133} & \textbf{0.023} & \textbf{0.027} & 0.031 & 0.037 & 0.046 & 0.017 & 0.010 \\
 & HM 3k & \textbf{0.065} & \textbf{0.070} & \textbf{0.109} & \textbf{0.025} & \textbf{0.038} & 0.027 & 0.029 & 0.038 & 0.017 & 0.020 \\
 & HM 5k & \textbf{0.060} & \textbf{0.058} & \textbf{0.094} & \textbf{0.027} & \textbf{0.057} & 0.026 & 0.021 & 0.037 & 0.016 & 0.026 \\
 & IndigoBM & \textbf{0.428} & \textbf{0.418} & \textbf{0.783} & \textbf{0.126} & \textbf{0.464} & 0.331 & 0.326 & 0.599 & 0.084 & 0.388 \\
\cmidrule(lr){2-12}
 & \textit{Average} & \textbf{0.438} & \textbf{0.487} & \textbf{0.605} & 0.242 & 0.316 & 0.389 & 0.410 & 0.502 & \textbf{0.250} & \textbf{0.331} \\
\midrule
\multirow{24}{*}{\rotatebox[origin=c]{90}{\textbf{Family 3}}} & FB-25 & \textbf{0.387} & \textbf{0.386} & \textbf{0.608} & \textbf{0.135} & 0.444 & 0.330 & 0.319 & 0.472 & 0.127 & \textbf{0.496} \\
 & FB-50 & \textbf{0.334} & \textbf{0.355} & \textbf{0.501} & \textbf{0.100} & 0.376 & 0.291 & 0.292 & 0.395 & 0.093 & \textbf{0.429} \\
 & FB-75 & \textbf{0.397} & \textbf{0.444} & \textbf{0.625} & 0.111 & 0.289 & 0.345 & 0.367 & 0.505 & \textbf{0.120} & \textbf{0.374} \\
 & FB-100 & \textbf{0.446} & \textbf{0.481} & \textbf{0.671} & \textbf{0.132} & 0.317 & 0.340 & 0.361 & 0.511 & 0.105 & \textbf{0.319} \\
 & WK-25 & \textbf{0.310} & \textbf{0.346} & \textbf{0.490} & \textbf{0.102} & 0.166 & 0.287 & 0.324 & 0.443 & 0.083 & \textbf{0.219} \\
 & WK-50 & \textbf{0.175} & \textbf{0.184} & \textbf{0.274} & \textbf{0.059} & 0.182 & 0.142 & 0.128 & 0.231 & 0.059 & \textbf{0.215} \\
 & WK-75 & \textbf{0.386} & \textbf{0.472} & \textbf{0.597} & 0.069 & \textbf{0.134} & 0.329 & 0.381 & 0.537 & \textbf{0.070} & 0.126 \\
 & WK-100 & \textbf{0.178} & \textbf{0.263} & \textbf{0.242} & \textbf{0.009} & 0.019 & 0.119 & 0.151 & 0.201 & 0.006 & \textbf{0.020} \\
 & NL-0 & \textbf{0.364} & \textbf{0.431} & \textbf{0.656} & \textbf{0.132} & 0.158 & 0.269 & 0.313 & 0.440 & 0.121 & \textbf{0.163} \\
 & NL-25 & \textbf{0.399} & \textbf{0.390} & \textbf{0.769} & 0.115 & 0.097 & 0.332 & 0.350 & 0.559 & \textbf{0.144} & \textbf{0.154} \\
 & NL-50 & \textbf{0.394} & \textbf{0.437} & \textbf{0.757} & 0.086 & 0.081 & 0.334 & 0.394 & 0.577 & \textbf{0.108} & \textbf{0.110} \\
 & NL-75 & \textbf{0.355} & \textbf{0.394} & \textbf{0.626} & 0.126 & 0.137 & 0.310 & 0.309 & 0.526 & \textbf{0.135} & \textbf{0.207} \\
 & NL-100 & \textbf{0.469} & \textbf{0.511} & \textbf{0.767} & 0.140 & 0.214 & 0.405 & 0.442 & 0.634 & \textbf{0.142} & \textbf{0.300} \\
 & Metafam & \textbf{0.344} & -- & 0.178 & \textbf{0.510} & -- & 0.133 & -- & \textbf{0.189} & 0.078 & -- \\
 & FBNELL & \textbf{0.480} & \textbf{0.485} & \textbf{0.789} & \textbf{0.168} & 0.467 & 0.410 & 0.370 & 0.687 & 0.164 & \textbf{0.490} \\
 & Wiki MT1 tax & \textbf{0.240} & \textbf{0.281} & \textbf{0.486} & 0.004 & \textbf{0.003} & 0.203 & 0.161 & 0.423 & \textbf{0.007} & 0.002 \\
 & Wiki MT1 health & \textbf{0.297} & \textbf{0.446} & \textbf{0.480} & 0.055 & \textbf{0.301} & 0.121 & 0.298 & 0.115 & \textbf{0.056} & 0.268 \\
 & Wiki MT2 org & \textbf{0.084} & \textbf{0.079} & \textbf{0.221} & \textbf{0.035} & \textbf{0.002} & 0.038 & 0.032 & 0.147 & 0.033 & 0.001 \\
 & Wiki MT2 sci & \textbf{0.254} & \textbf{0.268} & \textbf{0.494} & \textbf{0.029} & 0.002 & 0.212 & 0.167 & 0.442 & 0.026 & \textbf{0.004} \\
 & Wiki MT3 art & \textbf{0.251} & \textbf{0.268} & \textbf{0.408} & \textbf{0.115} & 0.060 & 0.150 & 0.160 & 0.259 & 0.049 & \textbf{0.062} \\
 & Wiki MT3 infra & \textbf{0.596} & \textbf{0.662} & \textbf{0.812} & 0.277 & 0.601 & 0.475 & 0.530 & 0.468 & \textbf{0.377} & \textbf{0.629} \\
 & Wiki MT4 sci & \textbf{0.293} & \textbf{0.326} & \textbf{0.434} & 0.011 & 0.254 & 0.174 & 0.171 & 0.335 & \textbf{0.013} & \textbf{0.349} \\
 & Wiki MT4 health & \textbf{0.525} & \textbf{0.566} & \textbf{0.653} & \textbf{0.250} & \textbf{0.412} & 0.284 & 0.314 & 0.369 & 0.094 & 0.210 \\
\cmidrule(lr){2-12}
 & \textit{Average} & \textbf{0.346} & \textbf{0.385} & \textbf{0.545} & \textbf{0.121} & 0.214 & 0.262 & 0.288 & 0.411 & 0.096 & \textbf{0.234} \\
\bottomrule
\end{tabular}
\end{adjustbox}
\caption{\textbf{Per-dataset scenario-stratified MRR for ULTRA and \ULTRArand (ULTRA with the relation encoder, entity encoder, and relation-graph initial features frozen at random initialization; only the score head is trained).} 
$^{\dagger}$Used during KGFM pretraining; reported for completeness, not zero-shot. Empty scenarios (no test triples) are reported as ``--''.}
\label{tab:app_ultraablation_perdataset}
\end{table*}

\section{Seen-query distractor diagnostic}
\label{sec:app-hown}

For each SQUA test triple $(h,r,t)$ we measure the $D(h,r)$ distraction rate:
how often the model scores the held-out target $t$ no higher than an entity in
$D(h,r)$, the set of seen $r$-answers of $h$ in the inference graph. These are
exactly the entities the filtered evaluation protocol removes from the rank, so
a high distraction rate is invisible to filtered MRR yet directly reflects the
seen-query prior ranking the previously observed $r$-tails of $h$ above the
correct answer. We report it for the pre-trained ULTRA 3-graph checkpoint on
three benchmarks chosen to span the range of SQUA degradation relative to
aggregate MRR: WN18RRInductive v4 (near-parity), WDsinger (intermediate), and
DBpedia100k (severe). Scores are unfiltered with pessimistic tie handling.
\cref{tab:hown-rates} shows the distraction rate tracks this degradation: where
it stays near parity ($\approx 0.5$) SQUA MRR stays close to Orig
(WN18RRInductive v4), and where it is high SQUA MRR falls far below (WDsinger,
DBpedia100k). This is the seen-query side of the asymmetry of \cref{sec:arch}:
the seen-query prior distracts from the held-out target unless the target aligns
with the seen answers $D(h,r)$.

 \begin{table}[t]
  \centering
  \small
  \setlength{\tabcolsep}{4pt}
  \begin{adjustbox}{max width=\columnwidth}
  \begin{tabular}{@{}l r rr r@{}}
  \toprule
  \multirow{2}{*}{\textbf{Dataset}}
    & \multirow{2}{*}{$n_{\text{SQUA}}$} 
    & \multicolumn{2}{c}{\textbf{ULTRA MRR}}
    & \multirow{2}{*}{\textbf{$D(h,r)$ distraction}} \\
  \cmidrule(lr){3-4}
    & & Orig & SQUA & \\
  \midrule
  DBpedia100k        & 17{,}415 & 0.391 & 0.145 & 0.664 \\
  WDsinger           &  1{,}224 & 0.386 & 0.163 & 0.916 \\
  WN18RRInductive v4 &  1{,}458 & 0.635 & 0.559 & 0.515 \\
  \bottomrule
  \end{tabular}
  \end{adjustbox}
  \caption{\textbf{On SQUA triples, SQUA MRR drops when the previously seen $r$-answers of $h$ ($D(h,r)$) outrank the held-out target $t$ instead of aligning with it}, with the $D(h,r)$
  distraction rate near parity only on the benchmark where SQUA MRR stays close to Orig. \textbf{ULTRA MRR} is the ULTRA aggregate MRR over all test triples (Orig) and its MRR on SQUA
  triples; the gap between them is the SQUA degradation these benchmarks span. The \textbf{$D(h,r)$ distraction rate} is the fraction of SQUA triples where some entity in $D(h,r)$ scores at
  least as high as $t$ under raw scores (pessimistic ties); the filtered protocol removes $D(h,r)$ from the rank, so a high distraction rate is invisible to filtered MRR.}
  \label{tab:hown-rates}
  \end{table}

\section{The query/answer asymmetry is not a node-degree artefact}
\label{sec:app-degree}

A natural objection to the query/answer asymmetry of \cref{sec:arch} is that it
might merely reflect target popularity: if UQSA targets are popular, high-degree
entities and SQUA targets unpopular, low-degree ones, a popularity bias alone
could reproduce the gap, with no half-link mechanism. To disentangle this, we
fix the node degree, binning test triples into gold-target-degree deciles (the
degree of $t$ in $\Gi$) and re-computing per-scenario MRR within each, on the 17
Family~2 benchmarks. The lower deciles are long-tail entities on which every
scenario scores near the floor, so we report the upper deciles.

 \cref{tab:app-degree} shows the asymmetry survives: UQSA exceeds SQUA in every
  reported decile, for example $0.511$ against $0.133$ at degree $34$--$55$. The
  gap is therefore not an artefact of target degree.

\begin{table}[t]
\centering
\small
\setlength{\tabcolsep}{6pt}
\begin{tabular}{@{}l l rrr@{}}
\toprule
\textbf{Decile} & \textbf{Degree} & Orig & SQUA & UQSA \\
\midrule
5  & 9--17        & 0.350 & 0.158 & 0.413 \\
7  & 34--55       & 0.377 & 0.133 & 0.511 \\
9  & 129--520     & 0.587 & 0.325 & 0.665 \\
10 & 523--6{,}411 & 0.839 & 0.721 & 0.859 \\
\bottomrule
\end{tabular}
\caption{\textbf{The query/answer asymmetry survives degree stratification: UQSA exceeds SQUA in every reported decile.} ULTRA MRR by gold-target-degree decile (degree in $\Gi$) on the 17
Family~2 benchmarks; the lower deciles are long-tail low-degree entities on which performance is near the floor and uninformative for this contrast, so the upper deciles are reported. Orig
is the aggregate MRR of ULTRA.}
\label{tab:app-degree}
\end{table}

\section{The query/answer asymmetry is not a relation-cardinality artefact}
\label{sec:app-cardinality}
A second possible confound for the query/answer asymmetry of \cref{sec:arch} is
relation cardinality: the SQUA $<$ UQSA gap could merely mirror 1-to-N queries being
harder than N-to-1 ones, with no half-link mechanism. To disentangle this, we
fix the relation cardinality, stratifying SQUA and UQSA MRR by the four
cardinality classes of~\citet{DBLP:conf/nips/BordesUGWY13} (1-to-1, 1-to-N,
N-to-1, N-to-N), pooled over the 17 Family~2 benchmarks.

\cref{tab:app-cardinality} shows the asymmetry survives: UQSA exceeds SQUA
within every cardinality class, for example $0.431$ against $0.061$ on 1-to-N.
The gap is therefore not an artefact of relation cardinality.

\begin{table}[t]
\centering
\small
\setlength{\tabcolsep}{8pt}
\begin{tabular}{@{}l rr r@{}}
\toprule
\textbf{Cardinality} & SQUA & UQSA & $\Delta$ \\
\midrule
Orig (all) & 0.140 & 0.572 & $+0.432$ \\
\midrule
1-to-1     & 0.494 & 0.692 & $+0.198$ \\
1-to-N     & 0.061 & 0.431 & $+0.370$ \\
N-to-1     & 0.206 & 0.540 & $+0.334$ \\
N-to-N     & 0.307 & 0.659 & $+0.352$ \\
\bottomrule
\end{tabular}
\caption{\textbf{Conditioning on relation cardinality leaves the query/answer asymmetry intact: UQSA exceeds SQUA within every cardinality class.} ULTRA MRR, SQUA versus UQSA, pooled over the 17 Family~2 benchmarks, stratified by the relation-cardinality categories of~\citet{DBLP:conf/nips/BordesUGWY13}. $\Delta = $ UQSA $-$ SQUA.}
\label{tab:app-cardinality}
\end{table}

\section{Per-benchmark fine-tuning results}
\label{sec:app-finetune}

We report results under a \emph{per-benchmark fine-tuning} protocol on
Family~3. The pre-trained checkpoint is fine-tuned on the training graph $G$ of
each benchmark individually, then evaluated by scoring its held-out test triples
$\Eu$ with message passing over its inference graph $\Gi$, exactly as in the
zero-shot setting; the zero-shot baseline is the same pre-trained checkpoint
applied to $\Gi$ without fine-tuning. For Family~3, $G$ is disjoint from $\Gi$ in
both entities and relation names, and only the weights change: the GNN and $\Gi$
do not, and $\Gi$ is still only used at inference. A fine-tuning gain on $\Gi$ therefore
reflects in-$\Gi$ signal that the generic pre-trained weights leave unused rather
than anything carried over from $G$, whose entities and relation names never
appear in $\Gi$.

\cref{tab:finetune_app_indr} gives the per-dataset fine-tuned MRR for ULTRA,
MOTIF, and TRIX on Family~3. We restrict the fine-tuning analysis to Family~3
because it is the only family on which fine-tuning stays in the inductive transfer setting
(\cref{sec:found}): there $G$ shares neither entities nor relation names with
$\Gi$. In Family~1, $G$ shares both with $\Gi$, and in Family~2 it shares the
relation names.

\begin{table*}[t]
\centering
\scriptsize
\setlength{\tabcolsep}{3pt}
\begin{adjustbox}{max width=\textwidth}
\begin{tabular}{@{}l ccccc ccccc ccccc@{}}
\toprule
\multirow{2}{*}{\textbf{Dataset}}
  & \multicolumn{5}{c}{\textbf{ULTRA}}
  & \multicolumn{5}{c}{\textbf{MOTIF}}
  & \multicolumn{5}{c}{\textbf{TRIX}} \\
\cmidrule(lr){2-6} \cmidrule(lr){7-11} \cmidrule(lr){12-16}
 & Orig & SQSA & UQSA & SQUA & UQUA & Orig & SQSA & UQSA & SQUA & UQUA & Orig & SQSA & UQSA & SQUA & UQUA \\
\midrule
FB-25 & \textbf{0.386} & 0.378 & \textbf{0.594} & \textbf{0.139} & \textbf{0.495} & 0.378 & 0.369 & 0.586 & 0.135 & 0.486 & 0.379 & \textbf{0.393} & 0.591 & 0.100 & 0.429 \\
FB-50 & 0.335 & 0.345 & 0.480 & \textbf{0.111} & \textbf{0.425} & \textbf{0.336} & 0.349 & 0.497 & 0.105 & 0.405 & 0.329 & \textbf{0.354} & \textbf{0.500} & 0.080 & 0.375 \\
FB-75 & \textbf{0.407} & \textbf{0.447} & 0.630 & \textbf{0.127} & \textbf{0.333} & 0.397 & 0.435 & 0.621 & 0.122 & 0.321 & 0.386 & 0.445 & \textbf{0.634} & 0.073 & 0.237 \\
FB-100 & \textbf{0.439} & 0.475 & 0.657 & \textbf{0.140} & 0.247 & 0.438 & 0.474 & 0.653 & 0.136 & \textbf{0.273} & 0.423 & \textbf{0.476} & \textbf{0.667} & 0.061 & 0.147 \\
WK-25 & 0.307 & 0.321 & 0.484 & \textbf{0.122} & \textbf{0.228} & \textbf{0.311} & \textbf{0.334} & 0.492 & 0.114 & 0.202 & 0.290 & 0.326 & \textbf{0.512} & 0.045 & 0.120 \\
WK-50 & \textbf{0.149} & \textbf{0.156} & 0.246 & 0.033 & \textbf{0.171} & 0.147 & 0.146 & 0.249 & \textbf{0.041} & 0.167 & 0.142 & 0.150 & \textbf{0.259} & 0.019 & 0.115 \\
WK-75 & \textbf{0.371} & 0.444 & \textbf{0.586} & \textbf{0.072} & 0.128 & 0.368 & 0.441 & 0.580 & 0.072 & \textbf{0.130} & 0.364 & \textbf{0.450} & 0.583 & 0.045 & 0.086 \\
WK-100 & 0.163 & 0.230 & 0.235 & 0.012 & 0.016 & 0.166 & 0.233 & 0.239 & \textbf{0.016} & \textbf{0.018} & \textbf{0.181} & \textbf{0.259} & \textbf{0.266} & 0.002 & 0.006 \\
NL-0 & \textbf{0.340} & 0.407 & 0.560 & \textbf{0.151} & \textbf{0.191} & 0.327 & \textbf{0.424} & 0.576 & 0.116 & 0.123 & 0.338 & 0.423 & \textbf{0.671} & 0.072 & 0.096 \\
NL-25 & \textbf{0.400} & \textbf{0.440} & 0.746 & 0.104 & \textbf{0.131} & 0.389 & 0.403 & 0.732 & \textbf{0.111} & 0.110 & 0.360 & 0.319 & \textbf{0.770} & 0.046 & 0.090 \\
NL-50 & 0.418 & 0.438 & 0.801 & 0.101 & \textbf{0.138} & \textbf{0.422} & \textbf{0.446} & 0.813 & \textbf{0.102} & 0.107 & 0.402 & 0.437 & \textbf{0.829} & 0.045 & 0.065 \\
NL-75 & \textbf{0.378} & \textbf{0.404} & 0.673 & \textbf{0.131} & \textbf{0.178} & 0.361 & 0.370 & \textbf{0.681} & 0.114 & 0.147 & 0.340 & 0.363 & 0.672 & 0.078 & 0.102 \\
NL-100 & \textbf{0.478} & \textbf{0.505} & 0.767 & \textbf{0.170} & \textbf{0.278} & 0.465 & 0.471 & 0.787 & 0.151 & 0.162 & 0.464 & 0.483 & \textbf{0.805} & 0.113 & 0.178 \\
Metafam & 0.999 & -- & 0.997 & \textbf{1.000} & -- & 0.998 & -- & \textbf{1.000} & 0.995 & -- & \textbf{1.000} & -- & \textbf{1.000} & \textbf{1.000} & -- \\
FBNELL & \textbf{0.490} & 0.466 & 0.814 & \textbf{0.187} & \textbf{0.527} & 0.479 & 0.475 & 0.793 & 0.170 & 0.474 & 0.477 & \textbf{0.507} & \textbf{0.829} & 0.121 & 0.354 \\
Wiki MT1 tax & 0.368 & \textbf{0.320} & 0.512 & 0.260 & \textbf{0.021} & \textbf{0.455} & 0.311 & \textbf{0.526} & \textbf{0.446} & 0.018 & 0.403 & 0.302 & 0.514 & 0.340 & 0.016 \\
Wiki MT1 health & 0.378 & 0.514 & 0.618 & 0.087 & 0.301 & \textbf{0.386} & \textbf{0.517} & 0.629 & \textbf{0.093} & \textbf{0.302} & 0.378 & 0.494 & \textbf{0.630} & 0.082 & 0.302 \\
Wiki MT2 org & 0.102 & 0.086 & 0.359 & 0.090 & 0.003 & \textbf{0.105} & \textbf{0.087} & \textbf{0.379} & \textbf{0.105} & \textbf{0.004} & 0.098 & 0.084 & 0.360 & 0.049 & 0.003 \\
Wiki MT2 sci & 0.323 & \textbf{0.310} & 0.531 & 0.149 & \textbf{0.017} & 0.321 & 0.308 & 0.523 & 0.153 & 0.016 & \textbf{0.336} & 0.290 & \textbf{0.541} & \textbf{0.184} & 0.009 \\
Wiki MT3 art & \textbf{0.314} & 0.295 & \textbf{0.498} & \textbf{0.166} & 0.194 & 0.314 & 0.298 & 0.495 & 0.165 & \textbf{0.206} & 0.270 & \textbf{0.307} & 0.493 & 0.047 & 0.103 \\
Wiki MT3 infra & 0.662 & 0.672 & 0.833 & 0.479 & 0.631 & \textbf{0.685} & \textbf{0.686} & \textbf{0.835} & \textbf{0.538} & \textbf{0.646} & 0.675 & 0.680 & 0.833 & 0.513 & 0.638 \\
Wiki MT4 sci & 0.310 & 0.339 & \textbf{0.454} & \textbf{0.038} & \textbf{0.406} & \textbf{0.312} & \textbf{0.344} & 0.454 & 0.030 & 0.400 & 0.308 & 0.344 & 0.453 & 0.016 & 0.252 \\
Wiki MT4 health & 0.698 & 0.782 & 0.687 & 0.418 & 0.440 & 0.701 & 0.783 & 0.691 & \textbf{0.426} & \textbf{0.450} & \textbf{0.703} & \textbf{0.790} & \textbf{0.714} & 0.388 & 0.430 \\
\bottomrule
\end{tabular}
\end{adjustbox}
\caption{\textbf{Per-dataset scenario-stratified MRR for ULTRA, MOTIF, and TRIX fine-tuned on Family 3.} Each model is fine-tuned individually on the training graph of each benchmark; in each cell,
bold marks the best of the three models. Empty scenarios (no test triples) are reported as ``--''.}
\label{tab:finetune_app_indr}
\end{table*}

\section{Comparison with FLOCK}
\label{sec:app-flock}

FLOCK \citep{DBLP:journals/corr/abs-2510-01510} replaces the deterministic
message passing of ULTRA, MOTIF, and TRIX with probabilistic random-walk
ensembles. Across the $54$ zero-shot graphs it averages $0.391$ MRR, marginally
above the strongest GNN-based model, TRIX at $0.387$ (the FLOCK average is
taken from the original paper). On FB15k237\_10 we apply our scenario-stratified
evaluation to FLOCK and TRIX (\cref{tab:flock_fb10}): they reach the same
overall MRR ($0.246$) and similar per-scenario MRR, but FLOCK needs roughly
$186\times$ the inference time of TRIX ($147$ versus $0.8$ minutes on a single
NVIDIA A100 40\,GB GPU with an Intel Xeon Platinum 8368 host). We therefore exclude FLOCK from the main comparison
(\cref{sec:found}).

 \begin{table}[h]
\centering
\small
\begin{tabular}{lcc}
\toprule
& \textbf{TRIX} & \textbf{FLOCK} \\
\midrule
Orig & \textbf{0.246} & \textbf{0.246} \\
SQSA & \textbf{0.127} & \textbf{0.127} \\
UQSA & \textbf{0.481} & 0.476 \\
SQUA & 0.014 & \textbf{0.018} \\
UQUA & 0.043 & \textbf{0.055} \\ 
\midrule
Time (min) & \textbf{0.79} & 147.18 \\
\bottomrule
\end{tabular}
\caption{\textbf{On FB15k237\_10, TRIX and FLOCK reach the same overall MRR, but FLOCK needs roughly $186\times$ the inference time.} Zero-shot MRR per scenario (top) and inference time in
minutes (bottom); bold marks the best in each row (highest MRR, lowest time). Time measured on a single NVIDIA A100 (40\,GB) GPU with an Intel Xeon Platinum 8368 host.}
\label{tab:flock_fb10} 
\end{table}

\section{Dataset statistics}
\label{sec:app-datasets}
\cref{tab:app_datasets_transd,tab:app_datasets_inde,tab:app_datasets_indr} list
the size statistics of the Family~1, Family~2, and Family~3 benchmarks, and
\cref{tab:app_quadrants} reports their per-benchmark half-link scenario
proportions, the breakdown behind \cref{tab:proportions}.

\begin{table*}[!t]
\centering
\caption{Family~1 datasets (16). Train, Valid, Test denote triples in the respective set.
Task: \emph{h/t} predicts both heads and tails; \emph{tails} predicts tails only.}
\label{tab:app_datasets_transd}
\small
\begin{adjustbox}{max width=\textwidth}
\begin{tabular}{@{}lrrrrrrc@{}}
\toprule
\textbf{Dataset} & \textbf{Entities} & \textbf{Rels} & \textbf{Train} & \textbf{Valid} & \textbf{Test} & \textbf{Task} \\
\midrule
CoDEx Small~\citep{codex}       & 2{,}034   & 42      & 32{,}888      & 1{,}827   & 1{,}828   & h/t   \\
WDsinger~\citep{dackgr}         & 10{,}282  & 135     & 16{,}142      & 2{,}163   & 2{,}203   & h/t   \\
FB15k237\_10~\citep{dackgr}     & 11{,}512  & 237     & 27{,}211      & 15{,}624  & 18{,}150  & tails \\
FB15k237\_20~\citep{dackgr}     & 13{,}166  & 237     & 54{,}423      & 16{,}963  & 19{,}776  & tails \\
FB15k237\_50~\citep{dackgr}     & 14{,}149  & 237     & 136{,}057     & 17{,}449  & 20{,}324  & tails \\
FB15k237~\citep{DBLP:conf/acl-cvsc/ToutanovaC15}       & 14{,}541  & 237     & 272{,}115     & 17{,}535  & 20{,}466  & h/t   \\
CoDEx Medium~\citep{codex}      & 17{,}050  & 51      & 185{,}584     & 10{,}310  & 10{,}311  & h/t   \\
NELL23k~\citep{dackgr}          & 22{,}925  & 200     & 25{,}445      & 4{,}961   & 4{,}952   & h/t   \\
WN18RR~\citep{wn18rr}           & 40{,}943  & 11      & 86{,}835      & 3{,}034   & 3{,}134   & h/t   \\
AristoV4~\citep{ssl_rp}         & 44{,}949  & 1{,}605 & 242{,}567     & 20{,}000  & 20{,}000  & h/t   \\
Hetionet~\citep{hetionet}       & 45{,}158  & 24      & 2{,}025{,}177 & 112{,}510 & 112{,}510 & h/t   \\
NELL995~\citep{nell995}         & 74{,}536  & 200     & 149{,}678     & 543       & 2{,}818   & h/t   \\
CoDEx Large~\citep{codex}       & 77{,}951  & 69      & 551{,}193     & 30{,}622  & 30{,}622  & h/t   \\
ConceptNet100k~\citep{cnet100k} & 78{,}334  & 34      & 100{,}000     & 1{,}200   & 1{,}200   & h/t   \\
DBpedia100k~\citep{dbp100k}     & 99{,}604  & 470     & 597{,}572     & 50{,}000  & 50{,}000  & h/t   \\
YAGO310~\citep{yago310}         & 123{,}182 & 37      & 1{,}079{,}040 & 5{,}000   & 5{,}000   & h/t   \\
\bottomrule
\end{tabular}
\end{adjustbox}
\end{table*}

\begin{table*}[!t]
\caption{Family~2 datasets (18). Triples denote the number of edges of the graph given at training, validation, or test. Valid and Test denote triples to be predicted in the respective graph.}
\label{tab:app_datasets_inde}
\small
\begin{adjustbox}{max width=\textwidth}
\begin{tabular}{@{}lrrrrrrrrr@{}}
\toprule
\multirow{2}{*}{\textbf{Dataset}} & \multirow{2}{*}{\textbf{Rels}}
  & \multicolumn{2}{c}{\textbf{Training Graph}}
  & \multicolumn{3}{c}{\textbf{Validation Graph}}
  & \multicolumn{3}{c}{\textbf{Test Graph}} \\
\cmidrule(l){3-4}\cmidrule(l){5-7}\cmidrule(l){8-10}
& & \textbf{Entities} & \textbf{Triples}
  & \textbf{Entities} & \textbf{Triples} & \textbf{Valid}
  & \textbf{Entities} & \textbf{Triples} & \textbf{Test} \\
\midrule
FB v1~\citep{DBLP:conf/icml/TeruDH20}        & 180 & 1{,}594  & 4{,}245   & 1{,}594  & 4{,}245   & 489    & 1{,}093  & 1{,}993   & 411    \\
FB v2~\citep{DBLP:conf/icml/TeruDH20}        & 200 & 2{,}608  & 9{,}739   & 2{,}608  & 9{,}739   & 1{,}166 & 1{,}660 & 4{,}145   & 947    \\
FB v3~\citep{DBLP:conf/icml/TeruDH20}        & 215 & 3{,}668  & 17{,}986  & 3{,}668  & 17{,}986  & 2{,}194 & 2{,}501 & 7{,}406   & 1{,}731 \\
FB v4~\citep{DBLP:conf/icml/TeruDH20}        & 219 & 4{,}707  & 27{,}203  & 4{,}707  & 27{,}203  & 3{,}352 & 3{,}051 & 11{,}714  & 2{,}840 \\
WN v1~\citep{DBLP:conf/icml/TeruDH20}        & 9   & 2{,}746  & 5{,}410   & 2{,}746  & 5{,}410   & 630    & 922     & 1{,}618   & 373    \\
WN v2~\citep{DBLP:conf/icml/TeruDH20}        & 10  & 6{,}954  & 15{,}262  & 6{,}954  & 15{,}262  & 1{,}838 & 2{,}757 & 4{,}011   & 852    \\
WN v3~\citep{DBLP:conf/icml/TeruDH20}        & 11  & 12{,}078 & 25{,}901  & 12{,}078 & 25{,}901  & 3{,}097 & 5{,}084 & 6{,}327   & 1{,}143 \\
WN v4~\citep{DBLP:conf/icml/TeruDH20}        & 9   & 3{,}861  & 7{,}940   & 3{,}861  & 7{,}940   & 934    & 7{,}084 & 12{,}334  & 2{,}823 \\
NELL v1~\citep{DBLP:conf/icml/TeruDH20}      & 14  & 3{,}103  & 4{,}687   & 3{,}103  & 4{,}687   & 414    & 225     & 833      & 201    \\
NELL v2~\citep{DBLP:conf/icml/TeruDH20}      & 88  & 2{,}564  & 8{,}219   & 2{,}564  & 8{,}219   & 922    & 2{,}086 & 4{,}586   & 935    \\
NELL v3~\citep{DBLP:conf/icml/TeruDH20}      & 142 & 4{,}647  & 16{,}393  & 4{,}647  & 16{,}393  & 1{,}851 & 3{,}566 & 8{,}048   & 1{,}620 \\
NELL v4~\citep{DBLP:conf/icml/TeruDH20}      & 76  & 2{,}092  & 7{,}546   & 2{,}092  & 7{,}546   & 876    & 2{,}795 & 7{,}073   & 1{,}447 \\
ILPC Small~\citep{ilpc}    & 48  & 10{,}230 & 78{,}616  & 6{,}653  & 20{,}960  & 2{,}908 & 6{,}653 & 20{,}960  & 2{,}902 \\
ILPC Large~\citep{ilpc}    & 65  & 46{,}626 & 202{,}446 & 29{,}246 & 77{,}044  & 10{,}179 & 29{,}246 & 77{,}044 & 10{,}184 \\
HM 1k~\citep{ham_bm}       & 11  & 36{,}237 & 93{,}364  & 36{,}311 & 93{,}364  & 1{,}771 & 9{,}899 & 18{,}638  & 476    \\
HM 3k~\citep{ham_bm}       & 11  & 32{,}118 & 71{,}097  & 32{,}250 & 71{,}097  & 1{,}201 & 19{,}218 & 38{,}285 & 1{,}349 \\
HM 5k~\citep{ham_bm}       & 11  & 28{,}601 & 57{,}601  & 28{,}744 & 57{,}601  & 900    & 23{,}792 & 48{,}425 & 2{,}124 \\
IndigoBM~\citep{indigo}    & 229 & 12{,}721 & 121{,}601 & 12{,}797 & 121{,}601 & 14{,}121 & 14{,}775 & 250{,}195 & 14{,}904 \\
\bottomrule
\end{tabular}
\end{adjustbox}
\end{table*}

\begin{table*}[!t]
\caption{Family~3 datasets (23). Triples denote the number of edges of the graph given at training, validation, or test. Valid and Test denote triples to be predicted in the respective graph.}
\label{tab:app_datasets_indr}
\small
\begin{adjustbox}{max width=\textwidth}
\begin{tabular}{@{}lrrrrrrrrrrrr@{}}
\toprule
\multirow{2}{*}{\textbf{Dataset}}
  & \multicolumn{3}{c}{\textbf{Training Graph}}
  & \multicolumn{4}{c}{\textbf{Validation Graph}}
  & \multicolumn{4}{c}{\textbf{Test Graph}} \\
\cmidrule(l){2-4}\cmidrule(l){5-8}\cmidrule(l){9-12}
& \textbf{Entities} & \textbf{Rels} & \textbf{Triples}
& \textbf{Entities} & \textbf{Rels} & \textbf{Triples} & \textbf{Valid}
& \textbf{Entities} & \textbf{Rels} & \textbf{Triples} & \textbf{Test} \\
\midrule
FB-25~\citep{DBLP:conf/icml/LeeCW23}           & 5{,}190  & 163 & 91{,}571  & 4{,}097  & 216 & 17{,}147 & 5{,}716  & 4{,}097  & 216 & 17{,}147 & 5{,}716  \\
FB-50~\citep{DBLP:conf/icml/LeeCW23}           & 5{,}190  & 153 & 85{,}375  & 4{,}445  & 205 & 11{,}636 & 3{,}879  & 4{,}445  & 205 & 11{,}636 & 3{,}879  \\
FB-75~\citep{DBLP:conf/icml/LeeCW23}           & 4{,}659  & 134 & 62{,}809  & 2{,}792  & 186 & 9{,}316  & 3{,}106  & 2{,}792  & 186 & 9{,}316  & 3{,}106  \\
FB-100~\citep{DBLP:conf/icml/LeeCW23}          & 4{,}659  & 134 & 62{,}809  & 2{,}624  & 77  & 6{,}987  & 2{,}329  & 2{,}624  & 77  & 6{,}987  & 2{,}329  \\
WK-25~\citep{DBLP:conf/icml/LeeCW23}           & 12{,}659 & 47  & 41{,}873  & 3{,}228  & 74  & 3{,}391  & 1{,}130  & 3{,}228  & 74  & 3{,}391  & 1{,}131  \\
WK-50~\citep{DBLP:conf/icml/LeeCW23}           & 12{,}022 & 72  & 82{,}481  & 9{,}328  & 93  & 9{,}672  & 3{,}224  & 9{,}328  & 93  & 9{,}672  & 3{,}225  \\
WK-75~\citep{DBLP:conf/icml/LeeCW23}           & 6{,}853  & 52  & 28{,}741  & 2{,}722  & 65  & 3{,}430  & 1{,}143  & 2{,}722  & 65  & 3{,}430  & 1{,}144  \\
WK-100~\citep{DBLP:conf/icml/LeeCW23}          & 9{,}784  & 67  & 49{,}875  & 12{,}136 & 37  & 13{,}487 & 4{,}496  & 12{,}136 & 37  & 13{,}487 & 4{,}496  \\
NL-0~\citep{DBLP:conf/icml/LeeCW23}            & 1{,}814  & 134 & 7{,}796   & 2{,}026  & 112 & 2{,}287  & 763     & 2{,}026  & 112 & 2{,}287  & 763     \\
NL-25~\citep{DBLP:conf/icml/LeeCW23}           & 4{,}396  & 106 & 17{,}578  & 2{,}146  & 120 & 2{,}230  & 743     & 2{,}146  & 120 & 2{,}230  & 744     \\
NL-50~\citep{DBLP:conf/icml/LeeCW23}           & 4{,}396  & 106 & 17{,}578  & 2{,}335  & 119 & 2{,}576  & 859     & 2{,}335  & 119 & 2{,}576  & 859     \\
NL-75~\citep{DBLP:conf/icml/LeeCW23}           & 2{,}607  & 96  & 11{,}058  & 1{,}578  & 116 & 1{,}818  & 606     & 1{,}578  & 116 & 1{,}818  & 607     \\
NL-100~\citep{DBLP:conf/icml/LeeCW23}          & 1{,}258  & 55  & 7{,}832   & 1{,}709  & 53  & 2{,}378  & 793     & 1{,}709  & 53  & 2{,}378  & 793     \\
\midrule
Metafam~\citep{mtdea}          & 1{,}316  & 28  & 13{,}821  & 1{,}316  & 28  & 13{,}821 & 590     & 656     & 28  & 7{,}257  & 184     \\
FBNELL~\citep{mtdea}           & 4{,}636  & 100 & 10{,}275  & 4{,}636  & 100 & 10{,}275 & 1{,}055  & 4{,}752  & 183 & 10{,}685 & 597     \\
Wiki MT1 tax~\citep{mtdea}     & 10{,}000 & 10  & 17{,}178  & 10{,}000 & 10  & 17{,}178 & 1{,}908  & 10{,}000 & 9   & 16{,}526 & 1{,}834  \\
Wiki MT1 health~\citep{mtdea}  & 10{,}000 & 7   & 14{,}371  & 10{,}000 & 7   & 14{,}371 & 1{,}596  & 10{,}000 & 7   & 14{,}110 & 1{,}566  \\
Wiki MT2 org~\citep{mtdea}     & 10{,}000 & 10  & 23{,}233  & 10{,}000 & 10  & 23{,}233 & 2{,}581  & 10{,}000 & 11  & 21{,}976 & 2{,}441  \\
Wiki MT2 sci~\citep{mtdea}     & 10{,}000 & 16  & 16{,}471  & 10{,}000 & 16  & 16{,}471 & 1{,}830  & 10{,}000 & 16  & 14{,}852 & 1{,}650  \\
Wiki MT3 art~\citep{mtdea}     & 10{,}000 & 45  & 27{,}262  & 10{,}000 & 45  & 27{,}262 & 3{,}026  & 10{,}000 & 45  & 28{,}023 & 3{,}113  \\
Wiki MT3 infra~\citep{mtdea}   & 10{,}000 & 24  & 21{,}990  & 10{,}000 & 24  & 21{,}990 & 2{,}443  & 10{,}000 & 27  & 21{,}646 & 2{,}405  \\
Wiki MT4 sci~\citep{mtdea}     & 10{,}000 & 42  & 12{,}576  & 10{,}000 & 42  & 12{,}576 & 1{,}397  & 10{,}000 & 42  & 12{,}516 & 1{,}388  \\
Wiki MT4 health~\citep{mtdea}  & 10{,}000 & 21  & 15{,}539  & 10{,}000 & 21  & 15{,}539 & 1{,}725  & 10{,}000 & 20  & 15{,}337 & 1{,}703  \\
\bottomrule
\end{tabular}
\end{adjustbox}
\end{table*}

\begin{table*}[!t]
\centering
\caption{\textbf{Per-benchmark half-link scenario proportions (\%) range from $0\%$ to $98\%$ SQSA,} the split-driven composition discussed in \cref{sec:proportions}
These proportions are a property of the test split alone; they do not depend on any model or on zero-shot vs.\ fine-tuned evaluation. Each row sums to 100\%. $N$ = number of test triples. $^\dagger$Tail-only evaluation task (predicting $t$ only); SQUA\,$\neq$\,UQSA.}
\label{tab:app_quadrants}
\small
\setlength{\tabcolsep}{4pt}
\begin{adjustbox}{max width=\textwidth}
\begin{tabular}{@{}c l l rrrr r@{}}
\toprule
& \textbf{Dataset} & \textbf{Ver.} & \textbf{SQSA} & \textbf{UQSA} & \textbf{SQUA} & \textbf{UQUA} & \textbf{$N$} \\
\midrule
\multirow{16}{*}{\rotatebox[origin=c]{90}{\textbf{Family 1}}}
 & CoDEx Small & --- & 80.6 & 9.2 & 9.2 & 1.0 & 3{,}656 \\
 & WDsinger & --- & 28.1 & 27.8 & 27.8 & 16.3 & 4{,}406 \\
 & FB15k237\_10$^{\dagger}$ & --- & 29.7 & 41.8 & 19.3 & 9.2 & 18{,}150 \\
 & FB15k237\_20$^{\dagger}$ & --- & 41.1 & 36.0 & 17.2 & 5.7 & 19{,}776 \\
 & FB15k237\_50$^{\dagger}$ & --- & 57.6 & 27.9 & 11.7 & 2.8 & 20{,}324 \\
 & FB15k237 & --- & 68.0 & 15.2 & 15.2 & 1.6 & 40{,}932 \\
 & CoDEx Medium & --- & 66.3 & 16.3 & 16.3 & 1.1 & 20{,}622 \\
 & NELL23k & --- & 41.7 & 25.0 & 25.0 & 8.3 & 9{,}904 \\
 & WN18RR & --- & 22.1 & 32.8 & 32.8 & 12.3 & 6{,}268 \\
 & AristoV4 & --- & 61.6 & 17.8 & 17.8 & 2.8 & 40{,}000 \\
 & Hetionet & --- & 98.0 & 1.0 & 1.0 & 0.0 & 225{,}020 \\
 & NELL995 & --- & 0.0 & 38.2 & 38.2 & 23.6 & 5{,}636 \\
 & CoDEx Large & --- & 53.0 & 22.6 & 22.6 & 1.8 & 61{,}244 \\
 & ConceptNet100k & --- & 79.1 & 9.9 & 9.9 & 1.1 & 2{,}400 \\
 & DBpedia100k & --- & 60.2 & 17.4 & 17.4 & 5.0 & 100{,}000 \\
 & YAGO310 & --- & 82.2 & 8.4 & 8.4 & 1.0 & 10{,}000 \\
\midrule
\multirow{18}{*}{\rotatebox[origin=c]{90}{\textbf{Family 2}}}
 & FB & v1 & 22.1 & 30.0 & 30.0 & 17.9 & 822 \\
 & FB & v2 & 37.2 & 24.6 & 24.6 & 13.6 & 1{,}894 \\
 & FB & v3 & 37.0 & 25.0 & 25.0 & 13.0 & 3{,}462 \\
 & FB & v4 & 43.4 & 23.1 & 23.1 & 10.4 & 5{,}680 \\
 & WN & v1 & 35.6 & 23.5 & 23.5 & 17.4 & 746 \\
 & WN & v2 & 32.3 & 23.8 & 23.8 & 20.1 & 1{,}704 \\
 & WN & v3 & 15.0 & 33.6 & 33.6 & 17.8 & 2{,}286 \\
 & WN & v4 & 31.0 & 25.8 & 25.8 & 17.4 & 5{,}646 \\
 & NELL & v1 & 0.0 & 50.0 & 50.0 & 0.0 & 402 \\
 & NELL & v2 & 48.1 & 21.0 & 21.0 & 9.9 & 1{,}870 \\
 & NELL & v3 & 41.2 & 25.9 & 25.9 & 7.0 & 3{,}240 \\
 & NELL & v4 & 56.4 & 18.6 & 18.6 & 6.4 & 2{,}894 \\
 & ILPC Small & --- & 44.6 & 26.3 & 26.3 & 2.8 & 5{,}804 \\
 & ILPC Large & --- & 31.5 & 32.7 & 32.7 & 3.1 & 20{,}368 \\
 & HM & 1k & 15.7 & 39.0 & 39.0 & 6.3 & 952 \\
 & HM & 3k & 11.7 & 41.0 & 41.0 & 6.3 & 2{,}698 \\
 & HM & 5k & 10.0 & 42.0 & 42.0 & 6.0 & 4{,}248 \\
 & IndigoBM & --- & 72.3 & 13.3 & 13.3 & 1.1 & 29{,}808 \\
\midrule
\multirow{23}{*}{\rotatebox[origin=c]{90}{\textbf{Family 3}}}
 & FB-25 & --- & 51.5 & 18.9 & 18.9 & 10.7 & 11{,}432 \\
 & FB-50 & --- & 40.5 & 22.3 & 22.3 & 14.9 & 7{,}758 \\
 & FB-75 & --- & 48.0 & 20.9 & 20.9 & 10.2 & 6{,}212 \\
 & FB-100 & --- & 60.0 & 18.2 & 18.2 & 3.6 & 4{,}658 \\
 & WK-25 & --- & 43.3 & 25.5 & 25.5 & 5.7 & 2{,}262 \\
 & WK-50 & --- & 41.5 & 26.2 & 26.2 & 6.1 & 6{,}450 \\
 & WK-75 & --- & 44.7 & 25.3 & 25.3 & 4.7 & 2{,}288 \\
 & WK-100 & --- & 43.0 & 25.6 & 25.6 & 5.8 & 8{,}992 \\
 & NL-0 & --- & 22.7 & 30.5 & 30.5 & 16.3 & 1{,}526 \\
 & NL-25 & --- & 20.0 & 35.3 & 35.3 & 9.4 & 1{,}488 \\
 & NL-50 & --- & 25.3 & 32.7 & 32.7 & 9.3 & 1{,}718 \\
 & NL-75 & --- & 32.2 & 28.3 & 28.3 & 11.2 & 1{,}214 \\
 & NL-100 & --- & 34.0 & 32.3 & 32.3 & 1.4 & 1{,}586 \\
 & Metafam & --- & 0.0 & 50.0 & 50.0 & 0.0 & 368 \\
 & FBNELL & --- & 37.0 & 27.6 & 27.6 & 7.8 & 1{,}194 \\
 & Wiki MT1 & tax & 8.2 & 44.2 & 44.2 & 3.4 & 3{,}668 \\
 & Wiki MT1 & health & 16.3 & 41.5 & 41.5 & 0.7 & 3{,}132 \\
 & Wiki MT2 & org & 87.9 & 5.9 & 5.9 & 0.3 & 4{,}882 \\
 & Wiki MT2 & sci & 19.9 & 38.3 & 38.3 & 3.5 & 3{,}300 \\
 & Wiki MT3 & art & 25.9 & 34.1 & 34.1 & 5.9 & 6{,}226 \\
 & Wiki MT3 & infra & 43.0 & 27.0 & 27.0 & 3.0 & 4{,}810 \\
 & Wiki MT4 & sci & 67.4 & 15.9 & 15.9 & 0.8 & 2{,}776 \\
 & Wiki MT4 & health & 64.2 & 17.2 & 17.2 & 1.4 & 3{,}406 \\
\bottomrule
\end{tabular}
\end{adjustbox}
\end{table*}

\end{document}